\documentclass[11pt]{article}

\usepackage[final]{acl}

\usepackage{times}
\usepackage{latexsym}

\usepackage[T1]{fontenc}

\usepackage[utf8]{inputenc}

\usepackage{microtype}

\usepackage{inconsolata}

\usepackage{graphicx}

\title{Does Accuracy Equal Evidence?\\Reasoning Faithfulness under KV Cache Compression}

\author{
 \textbf{Mengting Ai\textsuperscript{1}},
 \textbf{Jingrui He\textsuperscript{1}},
 \textbf{Yue Guo\textsuperscript{1}},
\\
 \textsuperscript{1}University of Illinois Urbana-Champaign\\
 \texttt{\{mai10, jingrui, yueg\}@illinois.edu}
}

\usepackage{multirow}      
\usepackage{caption}       
\usepackage[table]{xcolor} 
\usepackage{bbm}
\usepackage{graphicx} %
\usepackage{booktabs}
\usepackage{xspace}
\usepackage{amsmath} 
\usepackage[most]{tcolorbox}
\usepackage{pifont}
\usepackage{enumitem}
\usepackage{amssymb}

\providecommand{\better}[1]{\textcolor{red}{\textbf{#1}}}
\providecommand{\muted}[1]{\textcolor{gray}{#1}}
\providecommand{\best}[1]{\textbf{#1}}
\providecommand{\second}[1]{\underline{#1}}
\newcommand{\tightmidrule}{\noalign{\vskip 0.15ex}\hline\noalign{\vskip 0.15ex}}
\newcommand{\tightcmidrule}{\noalign{\vskip 0.05ex}\hline\noalign{\vskip 0.05ex}}

\begin{document}
\maketitle

\begin{abstract}
KV cache compression is commonly evaluated by final-answer accuracy, implicitly assuming that preserving the answer also preserves the reasoning that supports it. We test this assumption for large reasoning models and show that it can fail: under compression, correct answers and the validity of their visible supporting rationales can be preserved at different rates. We study this failure with a controlled fixed-trace replay protocol, which holds reasoning content fixed and isolates whether compression preserves usable information from an already available trace.
We evaluate ten token-eviction KV compression methods and one quantization method on three models across mathematical reasoning, scientific QA, clinical calculation, and long-context retrieval. We measure final accuracy, answer-chain consistency, and perturbation faithfulness. Across tasks, token-eviction methods can preserve competitive final-answer accuracy while substantially degrading chain support or perturbation faithfulness. We call this the \textbf{answer-evidence} gap. A coverage-preserving quantization control is substantially less affected, suggesting that the failure is tied less to KV memory reduction itself than to losing access to parts of the reasoning trace. Code is available at \url{https://github.com/famous-blue-raincoat/Safe_KV_Compress}.
\end{abstract}

\section{Introduction}

\begin{figure}
    \centering
    \includegraphics[trim={0.36cm 0.cm 0.38cm 0}, clip,width=1\linewidth]{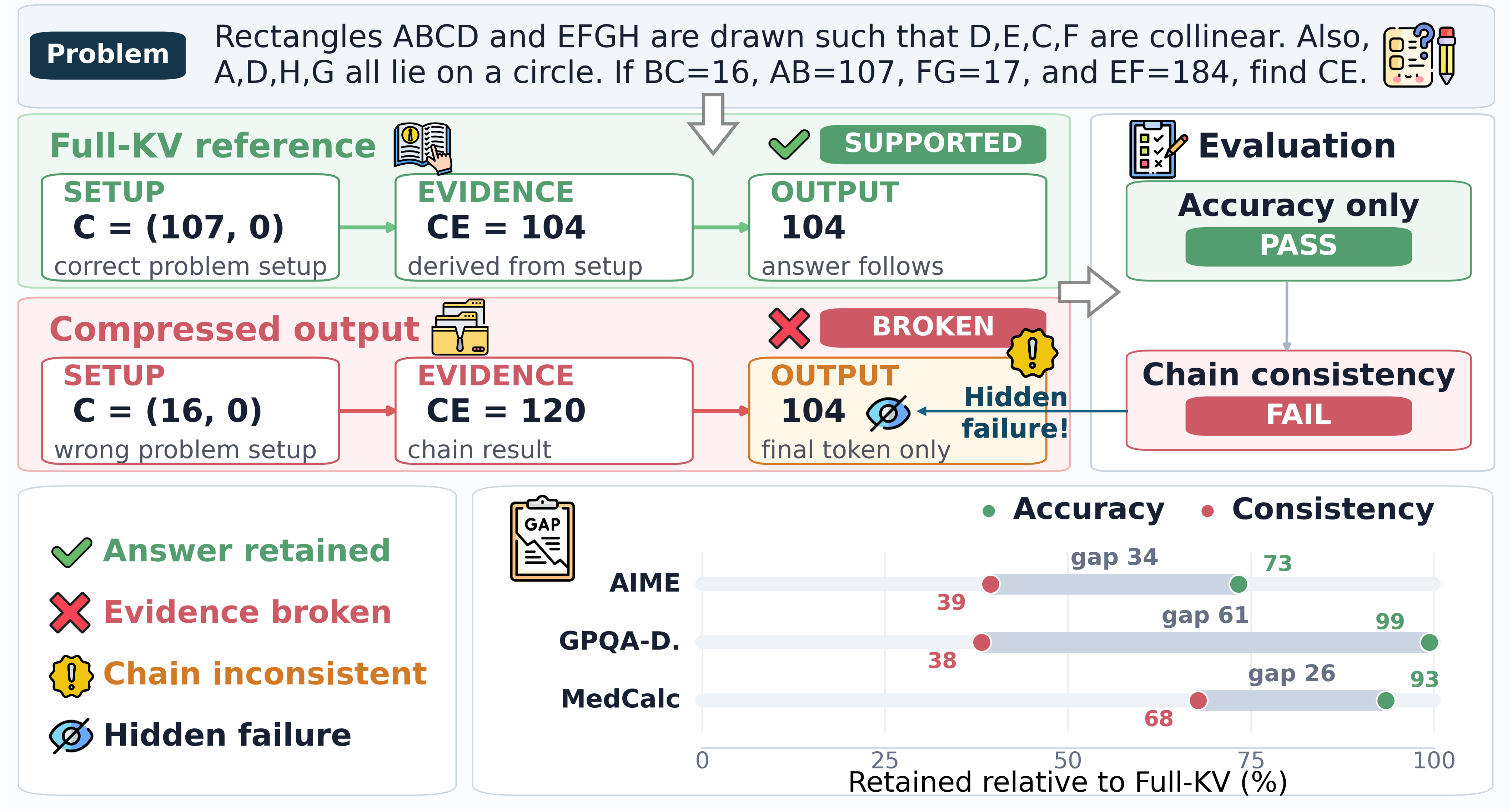}
    \vspace{-0.3in}
    \caption{Illustration of the answer--evidence gap under KV compression. The AIME example shows a compressed output that passes under accuracy, but fails chain consistency because its derivation uses unsupported evidence. The bottom plot shows that this gap appears across datasets: even when compressors retain much of the final-answer performance, the corresponding evidence support is much less preserved. 
}
    \label{fig:teaser}
    \vspace{-0.25in}
\end{figure}

Key-Value (KV) caching accelerates autoregressive decoding by storing the key and value states of previous tokens, avoiding repeated attention computation over the full prefix \citep{shutova2025cache}. This mechanism is increasingly critical as model scale and context length grow \citep{h2ozhang}, but it also introduces substantial memory overhead \citep{10.1145/3600006.3613165}. 
For large reasoning models (LRMs), which often generate thousands of intermediate reasoning tokens before the final answer, the KV cache can become a dominant inference-time memory bottleneck \citep{hao2026deltakvresidualbasedkvcache}.
This motivates a growing line of KV cache compression methods \citep{fu2025not,liu2026chunkkv}.

Most KV compression methods are evaluated by final-answer accuracy \citep{10.5555/3737916.3738638}. However, this metric is insufficient for LRMs, whose intermediate rationales are increasingly used to interpret, audit, and trust model outputs \citep{meek2025measuringchainofthoughtmonitorabilityfaithfulness}. Prior work on CoT faithfulness shows that plausible rationales may omit, distort, or post-hoc rationalize the evidence supporting an answer \citep{tutek-etal-2025-measuring,lanham2023measuringfaithfulnesschainofthoughtreasoning,barez2025chain}. 
KV compression may amplify this issue: a compressed cache can preserve enough information to recover the final answer while degrading the reasoning structure needed to justify or verify it. Thus, accuracy-only evaluation can produce false positives: a method is counted as successful even if the reasoning state does not support the answer,
which overestimates compression quality,
especially in settings where the rationale itself informs downstream decisions \citep{arcuschin2025chainofthoughtreasoningwildfaithful,NEURIPS2023_ed3fea90}.

We therefore study whether KV cache compression induces an \emph{answer--evidence gap}: final answers may remain correct even when the supporting reasoning becomes invalid, incomplete, or fragile to misleading intermediate claims. We frame this gap as an observable behavioral effect, rather than as evidence that the KV cache contains separate internal components for  ``answer'' and ``justification''. As Figure~\ref{fig:teaser} illustrates, answer preservation and evidence preservation can diverge sharply, making final accuracy an asymmetric diagnostic: accuracy collapse reveals compression damage, but preserved accuracy does not necessarily imply preserved evidential support.

To measure this gap, we use a controlled fixed-trace replay protocol \citep{kim2026caskcoreawareselectivekv,kim2026kvzip}. An uncompressed model first generates a reasoning trace, after which each compression method replays the same trace under compressed KV states. This isolates KV-retention behavior from rollout dynamics and allows us to ask whether compressed representations preserve not only the answer, but also the evidence supporting it.
We evaluate eleven widely used
compression methods across mathematical reasoning, scientific QA, clinical calculation, and long-context retrieval using three complementary metrics: final accuracy, answer--chain consistency, and perturbation faithfulness. 
Our results show that the answer--evidence gap is strongly task-dependent: on
AIME, GPQA-Diamond, and MedCalc, compressed models can maintain competitive accuracy while reasoning validity or faithfulness substantially degrades.
In contrast, on 
retrieval tasks such as RULER QA, where answers depend more directly on retaining exact supporting evidence, compression damage is more visible through accuracy collapse. To our best knowledge, this is the first study to evaluate KV compression through the lens of faithfulness.
Our contributions are:
\begin{itemize}[leftmargin=*,noitemsep,topsep=2pt]
    \item We identify an \emph{answer--evidence gap} as a false-positive failure mode of accuracy-only KV-compression: compressed models can preserve final answers while degrading the reasoning evidence supporting them.
    \item We show that the final--answer accuracy is an asymmetric diagnostic for compressed reasoning. While collapsed accuracy reveals compression damage, preserved accuracy can hide evidence loss, creating an illusion of competence where correct answer lacks supported evidence. 
    \item We provide a controlled measurement protocol combining final-answer accuracy, answer--chain consistency, and perturbation faithfulness, and show that these axes vary across budgets, compression methods, tasks, and model families.
\end{itemize}

\section{Problem Setting}
\label{sec:problem_setting}

\noindent\textbf{Compressed reasoning traces.}
Given a reasoning prefix $x_{1:t}$, a Full-KV model stores token-indexed key and value states $K_{1:t}^{(\ell,h)}$ and $V_{1:t}^{(\ell,h)}$ for each layer $\ell$ and head $h$. A KV compression method replaces this full cache with a smaller representation by retaining, merging, or evicting token states \citep{10.5555/3737916.3738638}.
For long reasoning traces, compression can therefore change which earlier tokens remain usable during generation.

\noindent\textbf{Answer cues and evidential support.}
Reasoning traces contain heterogeneous functional content \citep{tian2026skipkvselectiveskippingkv}. Some tokens make the final answer recoverable, such as answer-like claims, conclusion-adjacent summaries, or cues near the final response. Other tokens make the answer auditable, such as intermediate equations, premise checks, case splits, retrieved facts, and verification steps. We refer to these as \emph{answer-level cues} and \emph{evidential support}. This distinction is behavioral rather than mechanistic: we do not assume that KV states factorize into separate answer and justification components. Our question is whether compression preserves these two functions to the same degree.

\noindent\textbf{Controlled Fixed-Trace Replay}
For each question, we first run the uncompressed model and collect its complete reasoning trace. 
Each compression method is then evaluated on the same prompt and textual trace, but with its own compressed KV states. 
This controls for reasoning-content availability: all methods observe the same trace, so differences reflect how well each compressed KV representation preserves usable information from reasoning content that was already present. Thus, fixed-trace replay is a component-level diagnostic of KV retention rather than a deployment benchmark of end-to-end compressed generation \citep{ICLR2024_06a52a54,NEURIPS2022_6f1d43d5}. We also tested end-to-end results provided in Table \ref{tab:e2e-sanity} to show that the pattern still exists.

Concretely, given a question $q$ and ground-truth answer $y$, Full-KV generation first produces \texttt{<think>} $r$ \texttt{</think>} followed by an answer segment. We then prefill $q\,\Vert\,$\texttt{<think>} $r$ \texttt{</think>}, protect the question cache, compress only the cache corresponding to $r$, and resume decoding immediately after \texttt{</think>}. Thus, $q$ and the textual reasoning prefix $r$ are fixed across methods, whereas the post-\texttt{</think>} continuation, which contains the visible rationale and final answer, is newly generated from each compressed cache. Compression can therefore change whether the same available trace is usable for constructing a valid answer rationale, even though the replayed text is identical.

\section{Measuring Protocol}

Given a Full-KV reasoning trace, we evaluate whether compressed KV states preserve both answer recovery and evidential support. We use three complementary axes: (1) final-answer accuracy, (2) answer--chain consistency, and (3) perturbation-based faithfulness.

\subsection{Final Accuracy and Answer--Chain Consistency}
\label{sec:answer-chain}
 
Final-answer accuracy measures whether method $m$ outputs the ground-truth answer.
However, accuracy alone does not indicate whether the reasoning supports the answer. We therefore use an LLM-as-a-judge \citep{gu2025surveyllmasajudge} to separately label final correctness
and reasoning status.
We define wrong-chain correct answers as correct final answers whose reasoning is not fully correct (\textbf{R}easoning \textbf{W}rong \textbf{A}nswer \textbf{C}orrect):
A chain is treated as \emph{not fully correct} if, for example, it uses an incorrect setup, applies an invalid formula or theorem, makes an unsupported algebraic or logical transition, contradicts its own final answer, or simply hallucinates the answer without justification.

We use $\mathrm{RWAC}$ as the numerator and report two complementary metrics, with denominators given by all samples and answer-correct samples, respectively.
$\mathrm{RWAC}_{\mathrm{total}}$ measures benchmark-level prevalence: how often a method produces a correct-looking but unsupported answer overall. $\mathrm{RWAC}_{\mathrm{correct}}$ measures conditional reliability: among answers that would pass under final-answer accuracy, how often the reasoning is unsupported. We report both because low $\mathrm{RWAC}_{\mathrm{total}}$ can come from low accuracy, whereas $\mathrm{RWAC}_{\mathrm{correct}}$ tests whether accuracy-positive outputs are valid.

We validate the LLM judge with an independent human auditor. The human auditor independently applies the
same binary reasoning criterion: \texttt{correct} versus \texttt{not fully correct}. The auditor is a graduate-level researcher familiar with mathematical,
scientific, and long-context reasoning tasks.  We compute Cohen's $\kappa$ \citep{mchugh2012interrater} under the same binary reasoning label used in the main metrics. 
On 400 stratified outputs, human--Claude agreement is $94.6\%$ ($\kappa=0.89$). Re-evaluation with Gemini~3.1 Flash-Lite yields $94.2\%$ Claude--Gemini agreement ($\kappa=0.88$) and $95.0\%$ human--Gemini agreement ($\kappa=0.91$). A second human independently labels a 50-output subset, reaching $96.0\%$ inter-annotator agreement ($\kappa=0.92$).
Details appear in Appendix~\ref{sec:judge}.

\subsection{Perturbation-Based Faithfulness}
\label{sec:perturbation}

Accuracy and answer--chain consistency evaluate the static output. To test whether compressed KV states remain robust to unsupported intermediate claims, we insert an incorrect answer $z_i \neq y_i$ into the shared reasoning trace and measure whether the model changes its final answer or adopts the injected error, following prior CoT faithfulness evaluations~\citep{tutek-etal-2025-measuring,lanham2023measuringfaithfulnesschainofthoughtreasoning,barez2025chain}.
For a trace of length $T$, we insert the perturbation at one of three positions: \textbf{start}, immediately after the question prompt; \textbf{middle}, after token $\lfloor T/2 \rfloor$ in a trace of length $T$; and \textbf{end}, immediately before the model transitions to the final answer. The perturbation is a task-formatted declarative claim of the wrong answer. See Appendix~\ref{sec:perturb_case} for an example.

Let $\hat{y}_{i,0}^m$ denote
the unperturbed answer and $\hat{y}_{i,p}^m$ denote the answer after inserting
the perturbation at position $p \in \mathcal{P}$, where
$\mathcal{P}=\{\mathrm{start},\mathrm{middle},\mathrm{end}\}$. We report
\begin{align*}
\mathrm{Fidelity}(m)
&=
\mathbb{E}_{i,p}
\left[
\mathbbm{1}[\hat{y}_{i,p}^m = \hat{y}_{i,0}^m]
\right], \\
\mathrm{BiasRate}(m)
&=
\mathbb{E}_{i,p}
\left[
\mathbbm{1}[\hat{y}_{i,p}^m = z_i]
\right].
\end{align*}

Fidelity measures answer stability under misleading intervention, while Bias Rate measures directed adoption of the injected error. Since the perturbation is an explicit answer-like claim, this metric is a controlled probe of robustness to unsupported reasoning content rather than a complete adversarial robustness benchmark.

\begin{table}[t]
\centering
\small
\caption{Coverage of evaluated KV cache compression methods. We group methods by their primary compression signal to highlight the diversity of baselines. 
}
\vspace{-0.1in}
\label{tab:baseline-taxonomy}
\resizebox{\columnwidth}{!}{%
\begin{tabular}{p{2.7cm} p{3.7cm} p{3.4cm}}
\toprule
\textbf{Family} & \textbf{Methods} & \textbf{Primary signal} \\
\midrule

\textbf{Recency/attention sink} &
StreamingLLM~\citep{xiao2024efficient} &
Attention sinks and recent tokens. \\

\textbf{Attention scoring} &
SnapKV~\citep{10.5555/3737916.3738638} &
Observation-window attention statistics. \\

\textbf{Head-wise allocat.} &
HeadKV~\citep{fu2025not}, AdaKV~\citep{feng2026adakv} &
Head-level importance and adaptive budget allocation. \\

\textbf{Layer-wise allocat.} &
PyramidKV~\citep{cai2025pyramidkv} &
Layer-adaptive cache budgeting. \\

\textbf{Semantic span} &
ChunkKV~\citep{liu2026chunkkv} &
Contiguous chunks rather than isolated tokens. \\

\textbf{Local utility} &
KNorm~\citep{devoto-etal-2024-simple}, TOVA~\citep{oren-etal-2024-transformers} &
Key norms or online attention utility. \\

\textbf{Redundancy/reasoning aware} &
LagKV~\citep{liang2025lagkvlagrelativeinformationkv}, R-KV~\citep{cai2026rkv} &
Cache redundancy, relative information, or reasoning-token retention. \\

\textbf{Quantization} &
KIVI~\citep{liu2024kivi}
&
Low-bit KV quantization without token eviction. \\

\bottomrule
\end{tabular}%
}
\vspace{-0.15in}
\end{table}

\section{Experimental Results}
\label{sec:results}

We evaluate 11 representative KV cache compression methods across reasoning and retrieval tasks. Our main finding is an \emph{answer--evidence gap}: compressed models can preserve final-answer accuracy while degrading the derivation, verification, or robustness needed to support the answer.

\subsection{Setup}

\textbf{Datasets.}
We evaluate four tasks spanning a spectrum from answer-recoverable reasoning to evidence-constrained retrieval. In answer-recoverable tasks, partial cues in the reasoning trace may suffice to recover the final answer even when the full derivation is incomplete. In evidence-constrained tasks, correctness depends more directly on retaining specific supporting evidence.  \textbf{AIME24/25/26} \citep{aime24,aime25,aime26} tests multi-step symbolic derivation with integer answers. \textbf{GPQA-Diamond} \citep{rein2024gpqa} tests graduate-level scientific reasoning in a multiple-choice format. \textbf{MedCalc-Bench} \citep{NEURIPS2024_99e81750} tests clinical-style numerical calculation, where valid reasoning requires extracting patient-specific variables and applying formulas. \textbf{RULER QA} \citep{hsieh2024ruler} tests long-context multi-hop retrieval over 32K-token contexts with distractors. See IF-Eval results in Appendix~\ref{sec:full-results}.

\noindent\textbf{Models and compression methods.}
Our primary model is Qwen3-8B \citep{yang2025qwen3technicalreport}. We validate the main patterns on DeepSeek-R1-Distill-Llama-8B \citep{Guo_2025} and Qwen3-30B-A3B, with results provided in \ref{sec:dpsk-qwen30b}. 
We evaluate a diverse set of KV compression baselines covering recency and attention sink, attention-based scoring, head- and layer-wise allocation, semantic-span retention, local utility scoring, redundancy-aware compression, reasoning-aware retention, and quantization-based controls. Table~\ref{tab:baseline-taxonomy} summarizes these method families.

\noindent\textbf{Compression budget.}
Unless otherwise stated, eviction-based methods use a 256-token retained-cache budget. This aggressive setting provides diagnostic separation among methods: larger budgets often made results cluster near Full-KV, making it harder to observe the difference; we further ablate budgets from 512 to 8,192 tokens in Section~\ref{sec:rq4-budget}. Appendix~\ref{sec:efficiency-profile} reports the corresponding measured KV-memory and decode-latency trade-offs.

\noindent\textbf{End-to-end scope check.}
Fixed-trace replay is our main diagnostic because it holds the reasoning trace fixed and isolates KV-retention behavior. To contextualize deployment-style generation, we additionally run end-to-end checks on AIME26 (Table~\ref{tab:e2e-sanity}) and GPQA-Diamond (Appendix Table~\ref{tab:gpqa-end-to-end}). These experiments serve as deployment-style scope checks:  online compression jointly changes trajectory construction and KV retention, whereas fixed-trace replay isolates the retention question.

\begin{table}[t]
\centering
\small
\setlength{\tabcolsep}{3pt}
\caption{
End-to-end results on AIME26. For a full version, please refer to Appendix \ref{sec:end-to-end}
}
\label{tab:e2e-sanity}
\vspace{-0.1in}
\begin{tabular}{lccc}
\toprule
\textbf{Method} & {\textbf{Acc.($\uparrow$)}}&\textbf{RWAC/crct. ($\downarrow$)} & { \textbf{Fid.($\uparrow$)}} \\
\midrule
Full-KV &  56.7 & 11.8 & 91.1 \\
AdaKV &  23.3 &14.3 & 35.6 \\
HeadKV &  26.7 & 0.0& 31.1 \\
SnapKV &  26.7 &0.0 & 30.0 \\
StreamingLLM &  6.7 &50.0& 6.7 \\
\bottomrule
\end{tabular}
\vspace{-0.2in}
\end{table}

\subsection{RQ1: When Does Accuracy Hide the Answer--evidence Gap?}
\label{sec:rq1-final-accuracy}

Table~\ref{tab:main-results} reports the main Qwen3-8B results on final-answer accuracy, answer--chain consistency (reasoning quality), and faithfulness. The answer--evidence gap appears mostly on the token-eviction methods: compressed methods often retain enough information to produce answers while losing the reasoning support needed to justify or defend them.

\noindent\textbf{No-think is more transparent.}
The no-think baseline helps separate lower capability from misleading reasoning. Without a long reasoning trace, it usually has lower final accuracy. Yet when it fails, it often fails transparently: it does not exhibit the same level of wrong-chain rationalization as many compressed reasoning runs. This matters because compressed reasoning can look more successful than no-think under final accuracy while being less auditable: the model gives a correct answer, but the visible rationale no longer supports it.

\noindent\textbf{Quantization preserves structural integrity over numerical precision.} We include KIVI-2bit as a KV quantization control. Unlike eviction-based methods, KIVI reduces the precision of cached states while preserving token coverage over the reasoning trace. Its results remain close to Full-KV across all metrics. This suggests that the answer-evidence gap is not an inevitable consequence of reducing KV memory. Rather, the failure is most pronounced for token eviction methods that remove parts of the reasoning trace. By discarding intermediate tokens, eviction methods risk breaking the dependency structure and verification steps required for valid derivations, which can lead the model into unsupported rationalization. Quantization, by contrast, retains the tokens in the reasoning process, which indicates that preserving broad coverage of evidential states matters more for reasoning faithfulness than preserving them at full precision.

\begin{table*}[t]
\centering
\renewcommand{\dbltopfraction}{0.95}    
\renewcommand{\textfraction}{0.05}
\large
\vspace{-0.12in}
\caption{Main results on Qwen3-8B. \best{Bold} and \second{underline} indicate the best and second-best compressed baselines with complete entries. \better{Red bold} indicates better than Original. No-think faithfulness cells are omitted because there is no reasoning prefix to perturb. Values are 3-seed means.}
\vspace{-0.12in}
\label{tab:main-results}

\resizebox{0.66\textwidth}{!}{%
\begin{tabular}{lccccc}
\toprule
\multirow{2}{*}{\textbf{Method}} & \multirow{2}{*}{\textbf{Final Acc. ($\uparrow$)}} & \multicolumn{2}{c}{\textbf{Reasoning Quality}} & \multicolumn{2}{c}{\textbf{Faithfulness}} \\
\cmidrule(lr){3-4} \cmidrule(lr){5-6}
& & \textbf{RWAC/total ($\downarrow$)} & \textbf{RWAC/correct ($\downarrow$)} & \textbf{Fidelity ($\uparrow$)} & \textbf{Bias Rate ($\downarrow$)} \\
\tightmidrule
\rowcolor{blue!10}\multicolumn{6}{l}{\textbf{AIME24-26: Math reasoning}} \\
\rowcolor{blue!4}Qwen3-8B-think & 67.4{\tiny \ensuremath{\pm} 1.7} & 3.0{\tiny \ensuremath{\pm} 1.3} & 4.3{\tiny \ensuremath{\pm} 1.8} & 92.8{\tiny \ensuremath{\pm} 2.2} & 0.4{\tiny \ensuremath{\pm} 0.0} \\
 \rowcolor{blue!4}\muted{Qwen3-8B-no-think} & \muted{18.9{\tiny \ensuremath{\pm} 1.9}} & \muted{1.1{\tiny \ensuremath{\pm} 1.9}} & \muted{4.8{\tiny \ensuremath{\pm} 8.2}} & \muted{--} & \muted{--} \\
\tightcmidrule
\rowcolor{blue!4}KIVI-2bit & {67.4{\tiny \ensuremath{\pm} 0.6}} & {3.0{\tiny \ensuremath{\pm} 0.7}} & {4.5{\tiny \ensuremath{\pm} 1.0}} & {90.7{\tiny \ensuremath{\pm} 0.4}} & {0.5{\tiny \ensuremath{\pm} 0.6}} \\
\tightmidrule
\rowcolor{blue!4}StreamingLLM & 20.0{\tiny \ensuremath{\pm} 1.9} & \second{4.8{\tiny \ensuremath{\pm} 1.3}} & \second{23.5{\tiny \ensuremath{\pm} 5.2}} & 22.8{\tiny \ensuremath{\pm} 2.2} & 9.0{\tiny \ensuremath{\pm} 1.2} \\
\rowcolor{blue!4}SnapKV & \best{50.0{\tiny \ensuremath{\pm} 4.8}} & 33.0{\tiny \ensuremath{\pm} 6.4} & 65.5{\tiny \ensuremath{\pm} 7.0} & \second{61.6{\tiny \ensuremath{\pm} 5.3}} & 2.2{\tiny \ensuremath{\pm} 0.7} \\
\rowcolor{blue!4}HeadKV & 48.1{\tiny \ensuremath{\pm} 2.6} & 28.1{\tiny \ensuremath{\pm} 4.2} & 58.5{\tiny \ensuremath{\pm} 8.3} & 61.1{\tiny \ensuremath{\pm} 1.3} & \second{1.5{\tiny \ensuremath{\pm} 0.4}} \\
\rowcolor{blue!4}AdaKV & 48.5{\tiny \ensuremath{\pm} 0.6} & 32.6{\tiny \ensuremath{\pm} 4.2} & 62.8{\tiny \ensuremath{\pm} 4.6} & \best{64.4{\tiny \ensuremath{\pm} 4.4}} & 2.2{\tiny \ensuremath{\pm} 0.4} \\
\rowcolor{blue!4}PyramidKV & 34.1{\tiny \ensuremath{\pm} 2.8} & 17.0{\tiny \ensuremath{\pm} 3.4} & 49.7{\tiny \ensuremath{\pm} 6.1} & 43.1{\tiny \ensuremath{\pm} 2.6} & 2.1{\tiny \ensuremath{\pm} 0.6} \\
\rowcolor{blue!4}ChunkKV & 45.2{\tiny \ensuremath{\pm} 1.7} & 25.6{\tiny \ensuremath{\pm} 2.9} & 61.7{\tiny \ensuremath{\pm} 6.7} & 51.6{\tiny \ensuremath{\pm} 7.2} & 3.8{\tiny \ensuremath{\pm} 1.5} \\
\rowcolor{blue!4}KNorm & 15.2{\tiny \ensuremath{\pm} 1.3} & \better{2.6{\tiny \ensuremath{\pm} 1.3}} & \best{16.9{\tiny \ensuremath{\pm} 8.1}} & 19.6{\tiny \ensuremath{\pm} 0.4} & \best{0.7{\tiny \ensuremath{\pm} 1.0}} \\
\rowcolor{blue!4}TOVA & 33.3{\tiny \ensuremath{\pm} 3.8} & 17.8{\tiny \ensuremath{\pm} 1.9} & 53.4{\tiny \ensuremath{\pm} 0.4} & 37.8{\tiny \ensuremath{\pm} 2.1} & 6.3{\tiny \ensuremath{\pm} 1.0} \\
\rowcolor{blue!4}LagKV & 36.3{\tiny \ensuremath{\pm} 3.2} & 17.4{\tiny \ensuremath{\pm} 2.8} & 48.3{\tiny \ensuremath{\pm} 4.4} & 43.0{\tiny \ensuremath{\pm} 4.3} & 4.0{\tiny \ensuremath{\pm} 1.2} \\
\rowcolor{blue!4}R-KV & \second{49.6{\tiny \ensuremath{\pm} 3.9}} & 26.3{\tiny \ensuremath{\pm} 5.5} & 58.7{\tiny \ensuremath{\pm} 6.3} & 57.7{\tiny \ensuremath{\pm} 0.4} & 2.5{\tiny \ensuremath{\pm} 0.2} \\
\tightmidrule
\rowcolor{green!10}\multicolumn{6}{l}{\textbf{GPQA-Diamond: General reasoning}} \\
\rowcolor{green!4}Qwen3-8B-think & 57.7{\tiny \ensuremath{\pm} 0.3} & 24.4{\tiny \ensuremath{\pm} 2.1} & 42.3{\tiny \ensuremath{\pm} 3.4} & 95.9{\tiny \ensuremath{\pm} 1.0} & 1.6{\tiny \ensuremath{\pm} 0.6} \\
\rowcolor{green!4}\muted{Qwen3-8B-no-think} & \muted{50.3{\tiny \ensuremath{\pm} 1.2}} & \muted{20.5{\tiny \ensuremath{\pm} 2.0}} & \muted{40.8{\tiny \ensuremath{\pm} 4.0}} & \muted{--} & \muted{--} \\
\tightcmidrule
\rowcolor{green!4}KIVI-2bit & {58.8{\tiny \ensuremath{\pm} 1.8}} & {32.8{\tiny \ensuremath{\pm} 2.7}} & {55.5{\tiny \ensuremath{\pm} 3.6}} & {92.9{\tiny \ensuremath{\pm} 1.8}} & {4.3{\tiny \ensuremath{\pm} 1.2}} \\
\tightmidrule
\rowcolor{green!4}StreamingLLM & 46.8{\tiny \ensuremath{\pm} 2.3} & \best{30.0{\tiny \ensuremath{\pm} 3.0}} & \best{64.2{\tiny \ensuremath{\pm} 3.4}} & 49.1{\tiny \ensuremath{\pm} 1.0} & 22.2{\tiny \ensuremath{\pm} 3.7} \\
\rowcolor{green!4}SnapKV & 56.7{\tiny \ensuremath{\pm} 2.1} & 43.6{\tiny \ensuremath{\pm} 2.9} & 76.9{\tiny \ensuremath{\pm} 4.5} & 82.7{\tiny \ensuremath{\pm} 1.9} & 10.3{\tiny \ensuremath{\pm} 0.7} \\
\rowcolor{green!4}HeadKV & 56.4{\tiny \ensuremath{\pm} 1.5} & 37.2{\tiny \ensuremath{\pm} 0.6} & \second{66.0{\tiny \ensuremath{\pm} 2.2}} & 82.7{\tiny \ensuremath{\pm} 0.7} & \second{9.6{\tiny \ensuremath{\pm} 0.9}} \\
\rowcolor{green!4}AdaKV & 56.7{\tiny \ensuremath{\pm} 1.3} & 40.6{\tiny \ensuremath{\pm} 3.9} & 70.6{\tiny \ensuremath{\pm} 5.0} & \best{83.7{\tiny \ensuremath{\pm} 2.3}} & \best{8.0{\tiny \ensuremath{\pm} 1.2}} \\
\rowcolor{green!4}PyramidKV & 56.9{\tiny \ensuremath{\pm} 1.9} & 38.2{\tiny \ensuremath{\pm} 1.8} & 67.3{\tiny \ensuremath{\pm} 5.1} & \second{82.8{\tiny \ensuremath{\pm} 1.9}} & 10.2{\tiny \ensuremath{\pm} 0.5} \\
\rowcolor{green!4}ChunkKV & 56.7{\tiny \ensuremath{\pm} 2.1} & 42.4{\tiny \ensuremath{\pm} 2.3} & 74.0{\tiny \ensuremath{\pm} 6.0} & 81.4{\tiny \ensuremath{\pm} 1.0} & 10.5{\tiny \ensuremath{\pm} 2.1} \\
\rowcolor{green!4}KNorm & 43.9{\tiny \ensuremath{\pm} 2.8} & \second{37.0{\tiny \ensuremath{\pm} 4.2}} & 84.0{\tiny \ensuremath{\pm} 8.8} & 57.9{\tiny \ensuremath{\pm} 1.9} & 14.4{\tiny \ensuremath{\pm} 1.3} \\
\rowcolor{green!4}TOVA & \second{57.4{\tiny \ensuremath{\pm} 2.0}} & 43.6{\tiny \ensuremath{\pm} 0.3} & 76.3{\tiny \ensuremath{\pm} 3.0} & 77.9{\tiny \ensuremath{\pm} 2.1} & 12.5{\tiny \ensuremath{\pm} 2.1} \\
\rowcolor{green!4}LagKV & \better{57.9{\tiny \ensuremath{\pm} 1.9}} & 52.2{\tiny \ensuremath{\pm} 1.8} & 90.2{\tiny \ensuremath{\pm} 5.5} & 81.5{\tiny \ensuremath{\pm} 0.6} & 14.1{\tiny \ensuremath{\pm} 0.3} \\
\rowcolor{green!4}R-KV & 55.4{\tiny \ensuremath{\pm} 1.8} & 52.4{\tiny \ensuremath{\pm} 3.8} & 88.6{\tiny \ensuremath{\pm} 9.3} & 80.8{\tiny \ensuremath{\pm} 1.2} & 10.2{\tiny \ensuremath{\pm} 2.5} \\
\tightmidrule
\rowcolor{orange!12}\multicolumn{6}{l}{\textbf{MedCalc: Medical calculation}} \\
\rowcolor{orange!5}Qwen3-8B-think & 43.8{\tiny \ensuremath{\pm} 1.0} & 2.4{\tiny \ensuremath{\pm} 1.6} & 6.2{\tiny \ensuremath{\pm} 4.1} & 96.2{\tiny \ensuremath{\pm} 0.2} & 0.7{\tiny \ensuremath{\pm} 0.2} \\
\rowcolor{orange!5}\muted{Qwen3-8B-no-think} & \muted{31.6{\tiny \ensuremath{\pm} 0.3}} & \muted{4.9{\tiny \ensuremath{\pm} 0.5}} & \muted{18.1{\tiny \ensuremath{\pm} 2.1}} & \muted{--} & \muted{--} \\
\tightcmidrule
\rowcolor{orange!5}KIVI-2bit & {44.0{\tiny \ensuremath{\pm} 1.2}} & {3.8{\tiny \ensuremath{\pm} 0.1}} & {8.6{\tiny \ensuremath{\pm} 0.2}} & {92.1{\tiny \ensuremath{\pm} 0.4}} & {4.0{\tiny \ensuremath{\pm} 0.2}} \\
\tightmidrule
\rowcolor{orange!5}StreamingLLM & 37.2{\tiny \ensuremath{\pm} 0.5} & \second{4.8{\tiny \ensuremath{\pm} 2.2}} & 14.2{\tiny \ensuremath{\pm} 6.4} & 41.8{\tiny \ensuremath{\pm} 0.7} & 14.0{\tiny \ensuremath{\pm} 0.8} \\
\rowcolor{orange!5}SnapKV & 40.0{\tiny \ensuremath{\pm} 1.1} & 4.9{\tiny \ensuremath{\pm} 2.4} & \best{13.3{\tiny \ensuremath{\pm} 6.6}} & 64.9{\tiny \ensuremath{\pm} 1.5} & 8.9{\tiny \ensuremath{\pm} 0.8} \\
\rowcolor{orange!5}HeadKV & 40.3{\tiny \ensuremath{\pm} 0.7} & 5.5{\tiny \ensuremath{\pm} 1.4} & 14.5{\tiny \ensuremath{\pm} 3.5} & 61.3{\tiny \ensuremath{\pm} 1.5} & 9.5{\tiny \ensuremath{\pm} 0.6} \\
\rowcolor{orange!5}AdaKV & \second{41.1{\tiny \ensuremath{\pm} 1.5}} & 5.6{\tiny \ensuremath{\pm} 0.4} & 14.3{\tiny \ensuremath{\pm} 1.0} & \best{68.9{\tiny \ensuremath{\pm} 1.4}} & 8.5{\tiny \ensuremath{\pm} 0.6} \\
\rowcolor{orange!5}PyramidKV & 38.9{\tiny \ensuremath{\pm} 1.3} & 7.1{\tiny \ensuremath{\pm} 1.3} & 19.4{\tiny \ensuremath{\pm} 2.7} & 55.2{\tiny \ensuremath{\pm} 1.2} & \second{6.3{\tiny \ensuremath{\pm} 0.6}} \\
\rowcolor{orange!5}ChunkKV & 39.5{\tiny \ensuremath{\pm} 0.6} & 6.9{\tiny \ensuremath{\pm} 0.6} & 18.3{\tiny \ensuremath{\pm} 2.4} & 60.7{\tiny \ensuremath{\pm} 0.8} & 9.9{\tiny \ensuremath{\pm} 0.6} \\
\rowcolor{orange!5}KNorm & 33.6{\tiny \ensuremath{\pm} 0.7} & \best{4.1{\tiny \ensuremath{\pm} 1.9}} & 13.8{\tiny \ensuremath{\pm} 5.6} & 34.9{\tiny \ensuremath{\pm} 1.2} & \best{3.8{\tiny \ensuremath{\pm} 0.7}} \\
\rowcolor{orange!5}TOVA & 37.5{\tiny \ensuremath{\pm} 0.9} & 6.9{\tiny \ensuremath{\pm} 0.9} & 21.3{\tiny \ensuremath{\pm} 2.2} & 46.7{\tiny \ensuremath{\pm} 0.2} & 13.0{\tiny \ensuremath{\pm} 1.0} \\
\rowcolor{orange!5}LagKV & \best{41.5{\tiny \ensuremath{\pm} 1.3}} & 5.4{\tiny \ensuremath{\pm} 1.5} & \second{13.4{\tiny \ensuremath{\pm} 3.2}} & \second{65.5{\tiny \ensuremath{\pm} 0.3}} & 6.4{\tiny \ensuremath{\pm} 0.5} \\
\rowcolor{orange!5}R-KV & 40.0{\tiny \ensuremath{\pm} 0.7} & 5.5{\tiny \ensuremath{\pm} 0.4} & 13.7{\tiny \ensuremath{\pm} 1.0} & 62.5{\tiny \ensuremath{\pm} 1.1} & 9.7{\tiny \ensuremath{\pm} 1.0} \\
\tightmidrule
\rowcolor{purple!10}\multicolumn{6}{l}{\textbf{RULER: Long-context retrieval QA}} \\
\rowcolor{purple!4}Qwen3-8B-think & 78.9{\tiny \ensuremath{\pm} 0.7} & 10.5{\tiny \ensuremath{\pm} 2.4} & 13.4{\tiny \ensuremath{\pm} 3.0} & 71.3{\tiny \ensuremath{\pm} 0.4} & 10.0{\tiny \ensuremath{\pm} 0.2} \\
\rowcolor{purple!4}\muted{Qwen3-8B-no-think} & \muted{47.9{\tiny \ensuremath{\pm} 0.2}} & \muted{9.6{\tiny \ensuremath{\pm} 0.1}} & \muted{20.0{\tiny \ensuremath{\pm} 0.2}} & \muted{--} & \muted{--} \\
\tightcmidrule
\rowcolor{purple!4}KIVI-2bit & {64.3{\tiny \ensuremath{\pm} 0.7}} & {8.4{\tiny \ensuremath{\pm} 1.0}} & {17.6{\tiny \ensuremath{\pm} 0.3}} & {63.9{\tiny \ensuremath{\pm} 1.1}} & {8.4{\tiny \ensuremath{\pm} 0.2}} \\
\tightcmidrule
\rowcolor{purple!4}StreamingLLM & 32.7{\tiny \ensuremath{\pm} 0.4} & 25.1{\tiny \ensuremath{\pm} 3.8} & 77.4{\tiny \ensuremath{\pm} 11.7} & 14.7{\tiny \ensuremath{\pm} 0.1} & \better{5.5{\tiny \ensuremath{\pm} 0.1}} \\
\rowcolor{purple!4}SnapKV & \second{59.5{\tiny \ensuremath{\pm} 0.7}} & 17.1{\tiny \ensuremath{\pm} 4.8} & \second{29.0{\tiny \ensuremath{\pm} 8.2}} & \second{52.8{\tiny \ensuremath{\pm} 0.3}} & 11.1{\tiny \ensuremath{\pm} 0.7} \\
\rowcolor{purple!4}HeadKV & \best{61.2{\tiny \ensuremath{\pm} 1.1}} & \best{14.9{\tiny \ensuremath{\pm} 4.1}} & \best{24.0{\tiny \ensuremath{\pm} 6.6}} & \best{53.5{\tiny \ensuremath{\pm} 0.2}} & 10.9{\tiny \ensuremath{\pm} 0.3} \\
\rowcolor{purple!4}AdaKV & 35.7{\tiny \ensuremath{\pm} 0.4} & 18.1{\tiny \ensuremath{\pm} 5.7} & 50.2{\tiny \ensuremath{\pm} 15.7} & 27.5{\tiny \ensuremath{\pm} 0.3} & 16.1{\tiny \ensuremath{\pm} 0.7} \\
\rowcolor{purple!4}PyramidKV & 34.9{\tiny \ensuremath{\pm} 1.8} & 18.6{\tiny \ensuremath{\pm} 5.9} & 55.4{\tiny \ensuremath{\pm} 17.5} & 17.3{\tiny \ensuremath{\pm} 0.2} & \better{3.8{\tiny \ensuremath{\pm} 0.2}} \\
\rowcolor{purple!4}ChunkKV & 31.8{\tiny \ensuremath{\pm} 1.1} & 17.3{\tiny \ensuremath{\pm} 4.9} & 55.9{\tiny \ensuremath{\pm} 15.9} & 20.9{\tiny \ensuremath{\pm} 0.3} & 18.0{\tiny \ensuremath{\pm} 0.4} \\
\rowcolor{purple!4}KNorm & 27.7{\tiny \ensuremath{\pm} 0.4} & 23.0{\tiny \ensuremath{\pm} 2.6} & 82.1{\tiny \ensuremath{\pm} 9.3} & 9.1{\tiny \ensuremath{\pm} 0.2} & \better{3.1{\tiny \ensuremath{\pm} 0.2}} \\
\rowcolor{purple!4}TOVA & 32.3{\tiny \ensuremath{\pm} 1.8} & \second{15.3{\tiny \ensuremath{\pm} 3.3}} & 49.5{\tiny \ensuremath{\pm} 10.8} & 26.4{\tiny \ensuremath{\pm} 0.2} & 16.3{\tiny \ensuremath{\pm} 0.8} \\
\rowcolor{purple!4}LagKV & 31.1{\tiny \ensuremath{\pm} 0.4} & 17.6{\tiny \ensuremath{\pm} 4.3} & 57.1{\tiny \ensuremath{\pm} 14.1} & 26.7{\tiny \ensuremath{\pm} 0.3} & 14.1{\tiny \ensuremath{\pm} 0.7} \\
\rowcolor{purple!4}R-KV & 31.5{\tiny \ensuremath{\pm} 0.4} & 16.9{\tiny \ensuremath{\pm} 4.3} & 53.4{\tiny \ensuremath{\pm} 14.3} & 29.0{\tiny \ensuremath{\pm} 0.8} & 15.3{\tiny \ensuremath{\pm} 0.8} \\
\bottomrule
\end{tabular}%
}
\vspace{-0.25in}
\end{table*}

\noindent\textbf{Answer-recoverable tasks hide the damage.}
AIME exposes the cleanest separation between answer recovery and proof support. The best compressed methods recover many final answers, yet most of their correct answers are paired with invalid or incomplete chains. GPQA shows a more fragile pattern. Because the task is multiple-choice, even the uncompressed model often selects the right option without a fully verifiable rationale by guessing the answers. Compression amplifies this weakness into a dominant failure mode among correct answers: LagKV can reach even slightly better accuracy on GPQA, however it fails drastically on chain consistency because the similarity-based eviction removes a lot of the useful information, and the model proceeds to output the prediction without any justification. MedCalc shifts the failure to robustness: formula-triggering information can survive, while verification anchors needed to reject misleading intermediate values do not.

\noindent\textbf{RULER is the boundary condition.}
RULER QA behaves differently. Because the answer depends on exact distant evidence, there is less opportunity to recover the answer from partial reasoning shortcuts. When compression removes the necessary evidence, accuracy itself collapses. This is not a separate failure mode so much as a control condition: when answer-recovery shortcuts are absent, final accuracy is informative.

\noindent\textbf{Takeaway.}
Accuracy is an asymmetric diagnostic. Accuracy collapse reveals compression damage, but preserved accuracy can hide an answer--evidence gap. The dangerous case is not when compressed models visibly fail; it is when \emph{they remain accurate while the evidential support for their answers becomes invalid or incomplete}.

\subsection{RQ2: Do Accuracy Rankings Select Evidence-preserving Methods?}
\label{sec:rq2-ranking}

Table~\ref{tab:rank-disagreement} compares the compressor rankings induced by final-answer accuracy with answer--chain consistency and perturbation fidelity. This asks whether final accuracy is a reliable proxy for faithful evidence-aware evaluation. We use Spearman's $\rho$ because our question is ranking-based rather than
linear-predictive. Accuracy and evidence-oriented metrics may relate non-linearly, but compression benchmarks are often used to rank and select methods. Spearman  correlation therefore directly tests whether an accuracy-based leaderboard would agree with an evidence-aware leaderboard.

\noindent\textbf{Accuracy can reward unsupported answers.}
On AIME/GPQA-Diamond, accuracy and reasoning quality move in opposite directions: methods that rank highly by final answer tend to rank poorly by chain validity. This supports a behavioral separation between answer recovery and derivation support. Accuracy-based selection can therefore favor methods that recover the answer while failing to preserve the evidence needed to make that answer auditable. Perturbation fidelity is more positively aligned with accuracy, partly because the perturbed prefixes expose additional answer- or claim-relevant cues that importance-based compressors can retain. In other words, a method may be robust to perturbation when the relevant tokens are made salient by the intervention, but still fail to preserve the distributed derivation structure required for a valid reasoning chain. 

\noindent\textbf{Same-answer cases still lose chain support.}
RWAC shows that compressed models can produce correct answers with invalid chains, but it does not by itself isolate whether compression degraded a previously supported answer. We therefore run a matched-answer analysis on AIME: we keep only examples where Full-KV and the compressed method produce the same correct final answer and the Full-KV reasoning is correct.
Table~\ref{tab:matched-answer-degradation} reports how often the compressed reasoning becomes not fully correct under this condition. The pattern is strong: coverage-preserving KIVI rarely degrades the chain, while eviction-based methods degrade the reasoning chain in a much more noticeable way.

\begin{table}[t]
\centering
\small
\caption{Rank correlation between final-answer accuracy and faithfulness-oriented metrics among compressed baselines. We report Spearman $\rho$ over methods. Negative correlations indicate that higher-accuracy methods tend to be worse under the metric.}
\label{tab:rank-disagreement}
\vspace{-0.1in}
\begin{tabular}{lcc}
\toprule
Dataset & Acc vs Reasoning & Acc vs Faith.\\
\midrule
AIME26 & $-0.95$ & $+0.92$  \\
GPQA-D. & $-0.20$ & $+0.45$ \\
MedCalc & $+0.31$ & $+0.94$  \\
RULER & $+0.61$ & $+0.63$ \\
\bottomrule
\end{tabular}
\vspace{-0.25in}
\end{table}

\noindent\textbf{Output-level failure signatures.}
We further audit AIME RWAC cases: outputs whose final answer is correct but whose reasoning chain is not. 
Table~\ref{tab:rwac-failure-causes} compares Full-KV with SnapKV, a representative high-accuracy compressed method on AIME. Both settings contain unsupported reasoning, but compression changes the failure signature. The most distinctive increase is answer-first rationalization: among SnapKV RWAC cases, almost all outputs reach the correct answer while constructing a post-hoc or unsupported derivation around it. Step discontinuities remain common, indicating that the chain often lacks a valid bridge from intermediate reasoning to the final answer. This supports the answer--evidence gap interpretation at the output level: the answer can survive compression even when the evidential path that justifies it is broken.
\begin{table}[t]
\centering
\small
\caption{
Matched-answer chain degradation on AIME. We condition on examples where
Full-KV and the compressed method produce the same correct final answer and
the Full-KV reasoning is judged correct. Each entry reports how often the
compressed chain becomes not fully correct under this matched condition.
}
\label{tab:matched-answer-degradation}
\vspace{-0.1in}
\setlength{\tabcolsep}{5pt}
\begin{tabular}{l r @{\quad\quad} l r}
\toprule
\textbf{Method} & \textbf{Degraded} & \textbf{Method} & \textbf{Degraded} \\
\midrule
KIVI-2bit     &  4.2\% & TOVA      & 50.6\%\\
KNorm         & 15.4\% & R-KV      & 57.7\%   \\
StreamingLLM  & 22.4\% & ChunkKV   & 59.4\% \\
LagKV         & 47.9\% & HeadKV    & 60.2\% \\
PyramidKV & 49.4\% & AdaKV     & 63.2\% \\
              &        & SnapKV    & 64.8\% \\
\bottomrule
\end{tabular}
\vspace{-0.1in}
\end{table}

\noindent\textbf{Takeaway.}
Accuracy-based evaluation can be misleading because it can select methods that preserve answers without preserving evidence. This matters especially in high-stakes settings such as clinical or medical decision support: a correct-looking output may be used only if its evidential basis can be checked. When compression preserves the answer but not the reasoning support, users may be unable to tell whether the answer is actually reliable, creating risks that final accuracy alone cannot reveal.

\subsection{RQ3: Which Retention Biases Make Inserted Claims Influential?}
\label{sec:position}

\begin{table}[t]
\centering
\small
\caption{Distribution of reasoning failure causes among RWAC cases on AIME. Counts are non-exclusive since the model could make several mistakes at once.}
\vspace{-0.1in}
\providecommand{\rwacpos}[1]{\textcolor{red!70!black}{\textbf{#1}}}
\providecommand{\rwacneg}[1]{\textcolor{green!45!black}{\textbf{#1}}}
\begin{tabular}{lccc}
\toprule
Failure cause & Full-KV & SnapKV & $\Delta$ \\
& ($n=36$) & ($n=89$) & \\
\midrule
Ans.-first rationalization & 58\% & \textbf{94\%} & \rwacpos{+36 } \\
Step discontinuity & 78\% & 71\% & \rwacneg{-7 } \\
Hallucinated theorem & 17\% & 24\% & \rwacpos{+7 } \\
Premise loss & 8\% & 13\% & \rwacpos{+5 } \\
Arithmetic error & 19\% & 15\% & \rwacneg{-5 } \\
Missing case split & 14\% & 8\% & \rwacneg{-6 } \\
Invalid algebraic transit. & 14\% & 16\% & \rwacpos{+2 } \\
\bottomrule
\end{tabular}
\label{tab:rwac-failure-causes}
\vspace{-0.25in}
\end{table}

\begin{figure}
    \centering
    \includegraphics[width=1\linewidth]{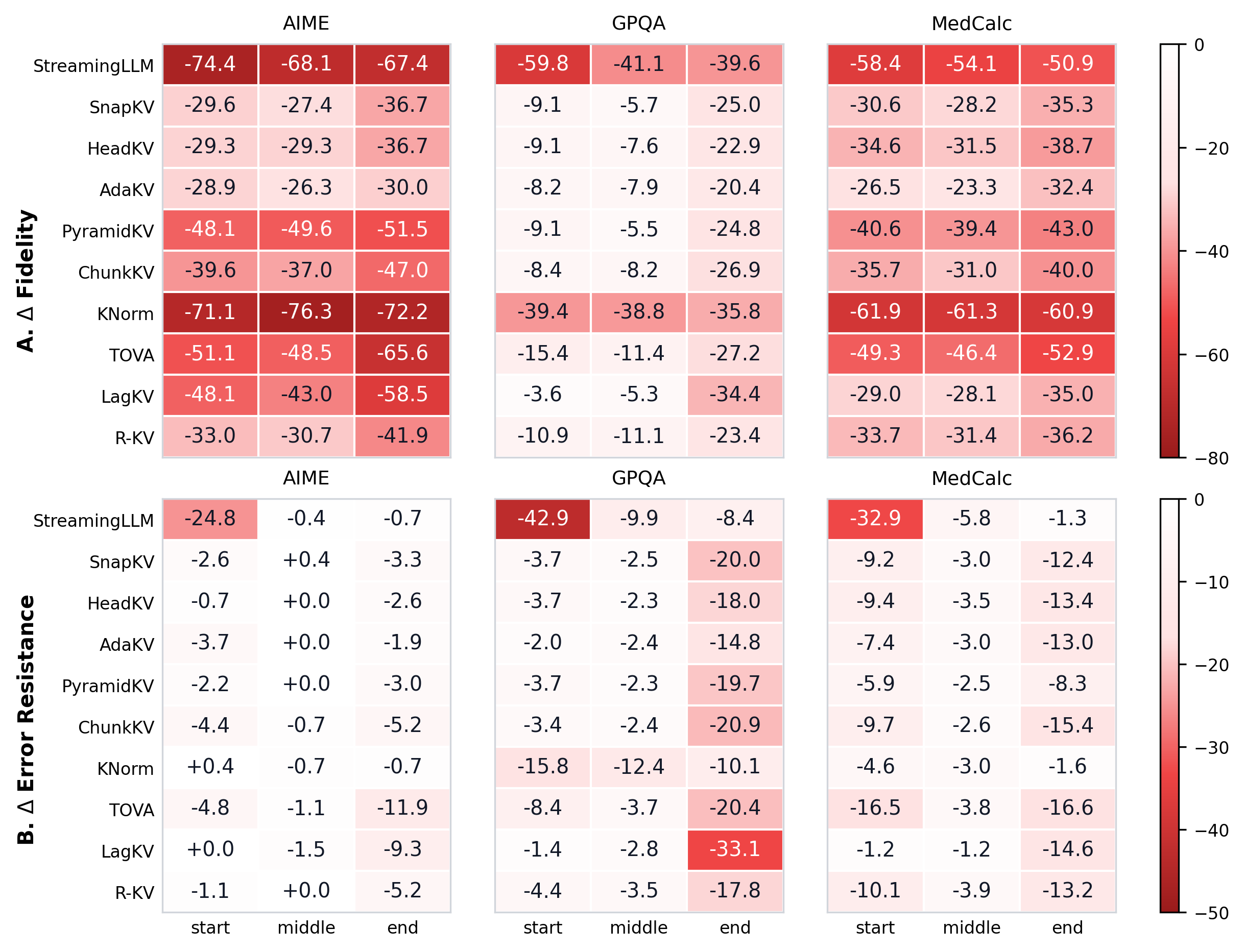}
    \vspace{-0.2in}
    \caption{
Position-conditioned changes in perturbation faithfulness under KV-cache compression. Each cell reports $\Delta=\mathrm{Score}_{\mathrm{compressed}}-\mathrm{Score}_{\mathrm{FullKV}}$. Negative values indicate degradation relative to matched Full-KV runs, and darker red indicates larger degradation. 
}
    \label{fig:position-intervention}
    \vspace{-0.25in}
\end{figure}

Figure~\ref{fig:position-intervention} shows that perturbation faithfulness is not degraded uniformly across compression methods. Instead, the failure pattern depends on each method's characteristics.

\noindent\textbf{Sink/window methods over-preserve early claims.}
StreamingLLM preserves attention sinks and a recent window. 
This prior retention can make start-position perturbations unusually influential: an early wrong answer may remain in the cache as an anchor while later corrective reasoning is evicted. The failure is therefore not just that compression removes information, but that it can preserve the wrong part of the prefix with high priority.

\begin{table}[t]
\centering
\small
\setlength{\tabcolsep}{5pt}
\caption{Wrong-chain span replacement results. A reasoning block in the trace is replaced by an in-domain wrong-chain block sampled from another seed. 
}
\vspace{-0.1in}
\label{tab:wrong-chain-replacement}
\begin{tabular}{llcc}
\toprule
\textbf{Dataset} & \textbf{Method} & \textbf{Fidelity ($\uparrow$)} & \textbf{Bias ($\downarrow$)} \\
\midrule
AIME26 & Qwen3-8B & 94.0 & 1.1 \\
 & StreamingLLM & 28.9 & 8.9\\
 & SnapKV & 53.0 & 5.6  \\
 & HeadKV & 61.4 & 7.8  \\
 & AdaKV & 66.3 & 3.4 \\
\midrule
GPQA-D. & Qwen3-8B & 96.5 & 2.5 \\
 & StreamingLLM & 60.2& 18.1 \\
 & SnapKV & 91.9  & 5.9 \\
 & HeadKV & 91.8 & 5.5 \\
 & AdaKV & 93.1  & 5.3  \\
\bottomrule
\end{tabular}
\vspace{-0.2in}
\end{table}

\noindent\textbf{Attention and head methods can over-preserve late conclusions.}
SnapKV, AdaKV, and HeadKV use attention or head-level signals to choose retained states. Late injected answers often look like summaries or conclusions, so they can be salient even when false. This mirrors suffix-style adversarial prompting, where appended content can strongly influence the final generation context \citep{zou2023universaltransferableadversarialattacks}. This indicates that salience is not trust: a conclusion-like token should not be retained without the upstream evidence that supports it.

\noindent\textbf{Redundancy and chunk methods preserve plausibility without verification.}
R-KV, LagKV, ChunkKV, and PyramidKV avoid the most extreme early-anchor failures, but they can still retain misleading content that is non-redundant, semantically coherent, or locally distinctive. Conversely, repeated verification steps may look redundant even when they are exactly what allows the model to reject a false conclusion. KNorm and TOVA show a related instability: large fidelity drops without always copying the injected answer, suggesting that low-level importance signals can destabilize reasoning even without direct error adoption.

\noindent\textbf{The gap persists under wrong-chain span replacement.}
The explicit wrong-answer injection is useful because it is controlled and easy to score, but it is not the only way a reasoning trace can become misleading. We therefore run an additional intervention in which a middle reasoning block is replaced by an in-domain wrong-chain block sampled from another seed. Table~\ref{tab:wrong-chain-replacement} shows that this evidence-level perturbation still exposes robustness degradation, especially on AIME. This supports the interpretation that the failure is not merely sensitivity to an answer cue; compressed traces can also be fragile to unsupported reasoning content.

\noindent\textbf{Takeaway.}
Position-conditioned perturbations expose why generic token importance is insufficient. Future reasoning compressors could explore preserving dependency and provenance structure: definitions, intermediate computations, and verification steps should be protected differently from unsupported answer-like conclusions.

\subsection{RQ4: Does More Cache Close the Answer--evidence Gap?}
\label{sec:rq4-budget}

\begin{figure}
    \centering
    \includegraphics[width=\linewidth]{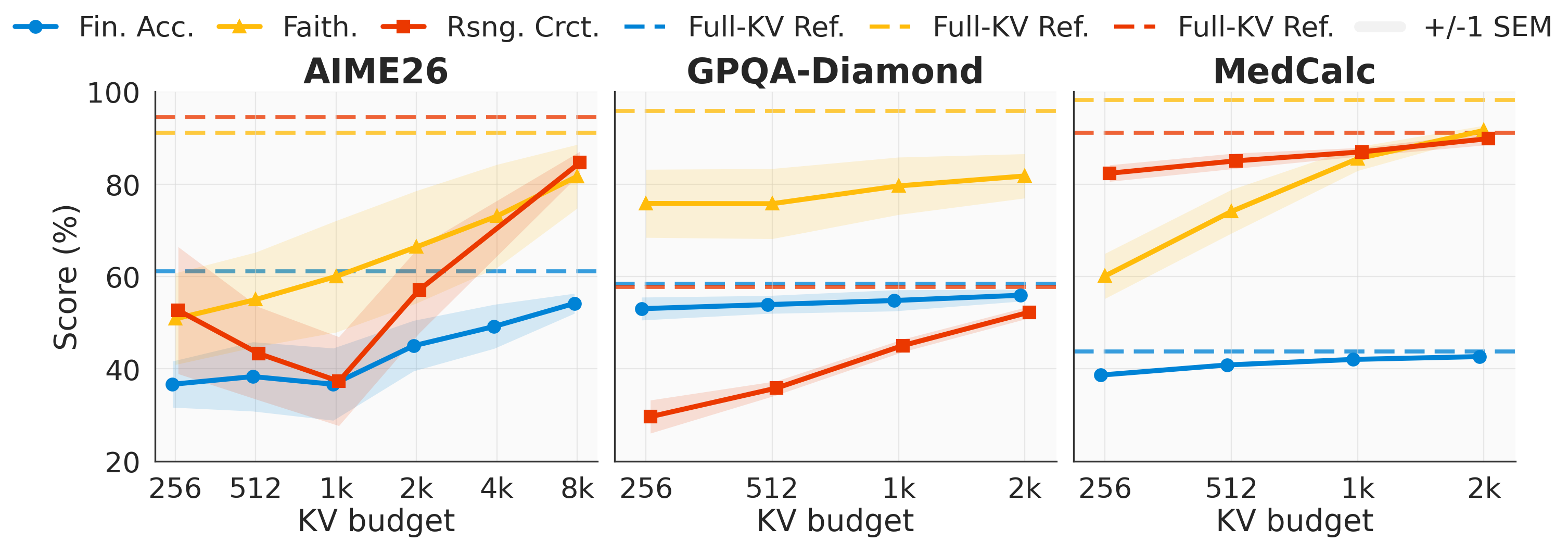}
    \vspace{-0.25in}
    \caption{Budget ablation on AIME26, GPQA-Diamond, and MedCalc (Qwen3-8B). On AIME, chain consistency (reasoning quality) remains degraded at all budgets while faithfulness recovers steadily; on GPQA/MedCalc, chain consistency recovers to near-baseline while faithfulness retains a persistent gap.
    }
  \vspace{-0.2in}
    \label{fig:budget-ablation}
\end{figure}

Figure~\ref{fig:budget-ablation} shows how compressed methods recover as the retained KV budget increases. Larger budgets generally improve the compressed model's accuracy, but the answer--chain consistency or perturbation fidelity still remains a large gap.

\noindent\textbf{Accuracy is the first metric to recover.}
Increasing the KV budget generally improves compressed methods across datasets and metrics. On AIME26, accuracy recovers substantially as the budget increases, and on GPQA-Diamond and MedCalc, it could almost recover to Full-KV performance. These trends show that giving the compressor more cache helps restore the answer prediction.

\noindent\textbf{The bottleneck depends on the task.}
For AIME, the persistent bottleneck is chain validity. Larger budgets improve fidelity, but a valid step-by-step derivation remains difficult because the proof evidence is distributed across many intermediate states. For GPQA and MedCalc, explicit answer support recovers more easily, but perturbation faithfulness remains below Full-KV: the model can produce a plausible rationale while still being vulnerable to a misleading injection.

\noindent\textbf{Takeaway.}
Larger budgets can improve final answers before closing the answer--evidence gap. It can recover enough information to answer, and sometimes enough to explain plausibly, while still failing to preserve the dependency and verification structure needed for robust reasoning.

\section{Related Work}

\noindent\textbf{KV cache compression.}
KV cache compression reduces inference memory by merging or evicting cached token states. Other methods also explore quantizing the KV cache \citep{NEURIPS2024_05d6b5b6,NEURIPS2024_028fcbcf,pmlr-v235-liu24bz}. For token eviction, early methods preserve heavy hitters, attention sinks, or recent windows \citep{h2ozhang,xiao2024efficient}; later work improves selection and allocation with observation-window attention \citep{10.5555/3737916.3738638}, pyramidal or layer-wise budgeting \citep{cai2025pyramidkv}, head-level allocation \citep{feng2026adakv}, semantic chunking \citep{liu2026chunkkv}, redundancy-aware compression, and online omission \citep{liang2025lagkvlagrelativeinformationkv,cai2026rkv}. Recent methods further organize the cache through head-level retrieval and reasoning-aware budgeting \citep{fu2025not}, composite tokens \citep{akulov2025kvcomposeefficientstructuredkv}, importance-diversity tradeoffs \citep{liu2026mixing}, dynamic retrieval with drift monitoring \citep{shi2026heterocachedynamicretrievalapproach}, sentence-level skipping for verbose CoT \citep{tian2026skipkvselectiveskippingkv}, residualized KV states \citep{hao2026deltakvresidualbasedkvcache}, or hierarchical memory routing \citep{liu2025mka}. 

\noindent\textbf{CoT faithfulness and monitorability.}
CoT can improve reasoning-task performance \citep{wei2023chainofthoughtpromptingelicitsreasoning,wang2023selfconsistencyimproveschainthought}, but generated rationales are not guaranteed to reveal the true driver of the answer. Prior work shows that CoTs can rationalize biased features, omit hint usage, or become unfaithful in realistic prompts \citep{NEURIPS2023_ed3fea90,lanham2023measuringfaithfulnesschainofthoughtreasoning,chen2025reasoningmodelsdontsay,arcuschin2025chainofthoughtreasoningwildfaithful}. Other work attempts to operationalize or improve faithfulness through symbolic solvers, causal and counterfactual tests, unlearning-based probes, instance-level benchmarks, and evidence-grounded verification \citep{meek2025measuringchainofthoughtmonitorabilityfaithfulness,tutek-etal-2025-measuring,shen2026faithcotbenchbenchmarkinginstancelevelfaithfulness}.

\noindent\textbf{Reasoning tokens, verbosity, and evidence.}
Pause tokens \citep{goyal2024thinkspeaktraininglanguage}, filler tokens \citep{pfau2024letsthinkdotdot}, and thinking tokens \citep{herel2024thinkingtokenslanguagemodeling} suggest that intermediate positions can provide additional computation even when their surface form is semantically uninformative. Recent methods further separate computation from readable rationales by reasoning in continuous states, or distilling System-2 traces into shorter System-1 behavior \citep{hao2025trainingcoconut,yu2024distilling21}. In parallel, concise and token-budgeted reasoning methods show that much shorter traces can often preserve final-answer accuracy through concise prompting, Chain of Draft \citep{xu2025chaindraftthinkingfaster}, compressed CoT \citep{lee2025llmscompresschainofthoughttoken}, sketch-style reasoning \citep{aytes2025sketchofthoughtefficientllmreasoning}, or learned short/long reasoning modes \citep{Renze_2024,kang2024c3otgeneratingshorterchainofthought,yi2025shorterbetterguidingreasoningmodels}.

\section{Conclusion}
\label{sec:conclusion}

We evaluate whether KV cache compression preserves not only final answers, but also the reasoning evidence that makes those answers auditable. Across eleven compression methods, four evaluation benchmarks, and three models, we find a consistent \textbf{answer--evidence gap}: compressed models can retain the correct final answer while producing invalid chains or becoming fragile to misleading context.
This makes final-answer accuracy an asymmetric diagnostic. When accuracy collapses, compression damage is visible; when accuracy remains high, it may hide evidence loss. 
Our results suggest that evaluation of KV compression for LRMs should move beyond accuracy-only leaderboards. Future compressors should be tested for answer--chain consistency and perturbation faithfulness, and future designs should preserve dependency, provenance, and verification structure rather than only tokens that appear salient for producing the final answer.

\clearpage
\newpage

\section*{Limitations}

Our study is diagnostic rather than exhaustive. First, the fixed-trace protocol isolates reasoning retention by holding the reasoning trace fixed, but it is not a full deployment simulation. In end-to-end compressed generation, compression can also change the reasoning trajectory, generation length, and overall capability.

Second, our evidence for the answer--evidence gap is behavioral. We show that answers, chain validity, and perturbation faithfulness can separate under compression, but we do not provide a causal token- or state-level account of which KV entries encode answer cues versus evidential support. Accordingly, our causal claims are intervention-level: varying the cache transformation while holding the model, prompt, textual prefix, and decoding configuration fixed can change the resulting behavior. We do not claim state-level semantic attribution or that the cache factorizes into distinct answer and evidence components.

Finally, answer--chain consistency relies on an LLM judge. We mitigate this with method-anonymized judging and human validation through different seeds, but exact RWAC values may vary with judge choice and prompt design.

\section*{Ethics Statement}
This work is primarily diagnostic and does not introduce a deployed system or collect personal user data. Its main ethical implication is cautionary. Accuracy-preserving KV compression may reduce the auditability of model reasoning, creating over-trust when users rely on rationales to assess correctness. This is especially relevant in high-stakes domains such as clinical calculation, scientific QA, and long-context evidence synthesis. We therefore recommend that compressed reasoning systems be evaluated with evidence-aware metrics, not only final-answer accuracy. We also note that our LLM-judge labels are intended for research evaluation and should not be treated as ground truth for real-world decisions without domain-expert review.

\section*{Acknowledgments}
This work is supported by National Science Foundation under Award No. IIS-2117902. We thank Xinrui He for her contributions to data annotation. The views and conclusions are those of the authors and should not be interpreted as representing the official policies of the funding agencies or the government.

\bibliography{custom}

\clearpage
\newpage

\appendix

\section{Additional Experiments}
\label{sec:full-results}

\begin{table*}[htbp]
\centering
\small
\caption{IF-Eval instruction-following results for Qwen3-8B. Values are 3-seed means. Strict prompt- and instruction-level accuracies are computed with the official IFEval checker. Intervention metrics are averaged over start/middle/end instruction-violation interventions. \textbf{Bold} and \underline{underline} indicate the best and second-best compressed baselines, respectively; \textit{italics} indicates better than Original.}
\label{tab:ifeval-qwen3-8b}
\begin{tabular}{lccccc}
\toprule
Method & Strict Prompt ($\uparrow$) & Strict Instr. ($\uparrow$) & Interv. Prompt ($\uparrow$) & Fidelity ($\uparrow$) & Bias/drop ($\downarrow$) \\
\midrule
Original & $85.8$ & $90.0$ & $80.5$ & $90.5$ & $5.1$ \\
\midrule
AdaKV & $\textit{85.8}$ & $\textit{90.5}$ & $\underline{\textit{82.0}}$ & $\textbf{\textit{91.4}}$ & $\textbf{\textit{3.9}}$ \\
ChunkKV & $\textit{86.1}$ & $\textit{90.6}$ & $\textit{81.8}$ & $\textit{90.8}$ & $\textit{4.3}$ \\
HeadKV & $85.5$ & $\textit{90.3}$ & $\textit{81.7}$ & $\textit{90.5}$ & $\textit{4.6}$ \\
KNorm & $84.8$ & $89.8$ & $\textit{82.0}$ & $\underline{\textit{91.1}}$ & $\underline{\textit{4.0}}$ \\
LagKV & $\textbf{\textit{86.4}}$ & $\textbf{\textit{90.8}}$ & $\textit{81.8}$ & $\textit{90.6}$ & $\textit{4.4}$ \\
PyramidKV & $\textit{85.8}$ & $\textit{90.4}$ & $\textbf{\textit{82.2}}$ & $\textit{90.9}$ & $\textit{4.2}$ \\
R-KV & $85.7$ & $\textit{90.4}$ & $\textit{81.9}$ & $\textit{90.6}$ & $\textit{4.4}$ \\
SnapKV & $\underline{\textit{86.4}}$ & $\underline{\textit{90.8}}$ & $\textit{82.0}$ & $\textit{90.6}$ & $\textit{4.4}$ \\
StreamingLLM & $85.6$ & $\textit{90.2}$ & $\textit{80.7}$ & $89.8$ & $5.4$ \\
TOVA & $84.9$ & $89.9$ & $\textit{82.0}$ & $\textit{90.8}$ & $\textit{4.2}$ \\
\bottomrule
\end{tabular}
\end{table*}

\begin{table*}[t]
\centering
\small
\setlength{\tabcolsep}{4pt}
\caption{Efficiency and faithfulness at the 256-token eviction budget. ``Evict. median/best'' summarizes perturbation fidelity across the token-eviction methods. Reasoning lengths and latency are median measurements; the length column also reports the 90th percentile.}
\label{tab:efficiency-profile}
\resizebox{\textwidth}{!}{%
\begin{tabular}{lrrrrrr}
\toprule
\textbf{Task} & \textbf{Reasoning tokens} & \textbf{Retained} & \textbf{KV cache (MiB)} & \textbf{Reduction} & \textbf{Fidelity: Full/Evict. med./best} & \textbf{Latency (ms/token)} \\
\midrule
AIME24--26 & 11,850 / 27,406 & 2.2\% & 2052.8 $\rightarrow$ 63.6 & 32.0$\times$ & 92.8 / 47.4 / 64.4 & 19.90 $\rightarrow$ 20.02 \\
GPQA-Diamond & 5,789 / 10,684 & 4.4\% & 1189.0 $\rightarrow$ 75.8 & 16.1$\times$ & 95.9 / 81.5 / 83.7 & 19.86 $\rightarrow$ 19.90 \\
MedCalc & 1,261 / 2,926 & 20.3\% & 358.9 $\rightarrow$ 169.2 & 2.2$\times$ & 96.2 / 61.0 / 68.9 & 19.79 $\rightarrow$ 19.83 \\
\bottomrule
\end{tabular}
}
\end{table*}

\subsection{Detailed Experimental Settings}

\paragraph{Implementation details.}
All experiments use HuggingFace Transformers with bfloat16 model weights and KV caches. Experiments were run on NVIDIA H100 80GB GPUs. The software stack uses \texttt{torch==2.6.0+cu124}, \texttt{transformers==4.57.6}, \texttt{accelerate==1.7.0}, and \texttt{kvpress==0.5.1}. 

\paragraph{Decoding settings.}
For thinking-mode Qwen3 and DeepSeek experiments, we use temperature $0.6$, top-$p$ $0.95$, top-$k$ $20$, min-$p$ $0.0$, and repetition penalty $1.0$. For the Qwen3-8B no-thinking ablation, we follow the Qwen non-thinking recommendation and use temperature $0.7$, top-$p$ $0.8$, top-$k$ $20$, and min-$p$ $0.0$. Fixed-reasoning traces are generated before compression. The teacher generation cap is 32,000 tokens for AIME, 12,000 for GPQA Diamond, and 4,096 for MedCalc.

\paragraph{Compressor implementations.}
SnapKV, StreamingLLM, TOVA, KNorm, LagKV, ChunkKV, AdaKV, and PyramidKV are implemented through kvpress \citep{devoto2025expectedattention}, with small wrappers that enforce prompt/suffix protection and exact target budgets.
R-KV and HeadKV are implemented following the corresponding public method designs.

\subsection{Efficiency Profile}
\label{sec:efficiency-profile}

Table~\ref{tab:efficiency-profile} places the diagnostic 256-token budget in practical context. The retained fraction is computed against the median reasoning length for each task. KV memory is measured for the full inference state before and after compression, and latency is decode time per generated token on the same H100 setup. The budget is intentionally aggressive: it yields large memory reductions on long-trace AIME and GPQA, while decode latency changes little because cache selection and model computation remain in the critical path. The resulting trade-off is therefore primarily memory versus behavioral faithfulness, rather than memory versus decoding speed.

\subsection{Additional Experimental Results}

\subsubsection{Results on IF-Eval}

IF-Eval provides a useful contrast to the reasoning-heavy tasks in the main text. 
Unlike AIME, GPQA-Diamond, or MedCalc, instruction-following evaluation usually does not require a long pre-answer derivation whose intermediate evidence must be audited. The output is judged primarily by whether it satisfies explicit surface constraints, such as formatting, length, or required content. As a result, there is less room for a wrong-chain-correct failure mode: the model either follows the instruction or it does not.

The results in Table~\ref{tab:ifeval-qwen3-8b} are consistent with this distinction. Most compressed methods remain close to the Full-KV model, and several slightly exceed it under the strict checker or intervention metrics. We do not interpret these small gains as evidence that compression generally improves instruction following; they are more likely due to the constrained nature of the task, mild denoising effects, or normal seed-level variation. Instead, the main takeaway is that the answer--evidence gap is not a universal degradation pattern across all tasks. It is most visible when the task requires preserving a long evidential chain, whereas constraint-following tasks with short, externally checkable outputs are less sensitive to the kind of reasoning-support loss studied in the main experiments.

\subsubsection{End-to-end results}
\label{sec:end-to-end}

In the main text, we report a compact end-to-end scope check to contextualize
deployment-style compressed generation. These results are not used as the main
evidence for the fixed-trace answer--evidence gap, because end-to-end generation
jointly varies trajectory construction, trace length, intermediate mistakes, and
KV retention.
Here, we provide the full results, including RWAC and perturbation metrics.
Unlike fixed-trace replay, end-to-end compressed generation is a deployable
setting: compression affects not only the retained representation of an
existing reasoning trace, but also the future trajectory, generation length,
intermediate mistakes, and final answer. We therefore treat end-to-end results
as a deployment-relevance check rather than as the primary diagnostic for the
answer--evidence gap.

Table~\ref{tab:aime26-qwen3-8b-end-to-end-rwfc} shows that deployment-style
compressed generation is also fragile: compressed end-to-end runs remain far
below Full-KV in final accuracy and perturbation fidelity even at a larger
2048-token budget.
Even with a larger budget, compressed end-to-end runs remain far below Full-KV in both
final accuracy and perturbation fidelity. This indicates that compression
damage is not merely an artifact of replaying a fixed teacher trace: in a
deployable generation setting, the model also becomes substantially less
stable under perturbation.

At the same time, the end-to-end setting changes the nature of the failure. In
fixed-trace replay, the same reasoning content is available to all methods, so
RWAC directly probes whether the compressed KV representation preserves the
evidence supporting an already generated trace. In end-to-end generation,
compression can instead alter the trajectory before such evidence is produced.
As a result, some methods show low RWAC/correct not because their correct
answers are reliably supported, but because only a small number of correct
answers survive. For this reason, we use fixed-trace replay as the main
faithfulness diagnostic and interpret end-to-end results primarily as evidence
that compressed generation remains fragile in deployment-like settings.

\begin{table*}[t]
\captionsetup{font=small}
\centering
\small
\setlength{\tabcolsep}{3pt}
\caption{Qwen3-8B AIME26 end-to-end KV-compression results at budget 2048, compared with our fixed-trace baseline at KV budget 256.
}
\label{tab:aime26-qwen3-8b-end-to-end-rwfc}
\begin{tabular}{llrrrrr}
\toprule
\textbf{Method} & \textbf{Setting} & \textbf{Final Acc. ($\uparrow$)} & \textbf{RWAC/total ($\downarrow$)} & \textbf{RWAC/correct ($\downarrow$)} & \textbf{Fidelity ($\uparrow$)} & \textbf{Bias Rate ($\downarrow$)} \\
\midrule
Qwen3-8B & Full-KV & 56.7 & 6.7 & 11.8 & 91.1 & 1.1 \\
\midrule
AdaKV & Fixed-trace (256) & 40.0 & 26.7 & 66.7 & 65.6 & 3.3 \\
 & End-to-end (2048) & 23.3 & 3.3 & 14.3 & 35.6 & 8.9 \\
\addlinespace[0.2ex]
HeadKV & Fixed-trace (256) & 46.7 & 30.0 & 64.3 & 61.1 & 2.2 \\
 & End-to-end (2048) & 26.7 & 0.0 & 0.0 & 31.1 & 4.4 \\
\addlinespace[0.2ex]
SnapKV & Fixed-trace (256) & 40.0 & 23.3 & 58.3 & 60.0 & 2.2 \\
 & End-to-end (2048) & 26.7 & 0.0 & 0.0 & 30.0 & 4.4 \\
\addlinespace[0.2ex]
StreamingLLM & Fixed-trace (256) & 20.0 & 0.0 & 0.0 & 16.7 & 11.1 \\
 & End-to-end (2048) & 6.7 & 3.3 & 50.0 & 6.7 & 11.1 \\
\bottomrule
\end{tabular}
\end{table*}

We additionally evaluate end-to-end compression on GPQA-Diamond at the same 2,048-token budget. Table~\ref{tab:gpqa-end-to-end} shows that the answer--evidence separation remains observable beyond AIME: compared with Full-KV, SnapKV, HeadKV, and AdaKV retain much of the answer accuracy but have higher RWAC/correct, while all eviction methods reduce perturbation fidelity. StreamingLLM shows a broader capability collapse. Because online compression also changes the generated trajectory, these results establish deployment relevance but are not used to localize retention failures.

\begin{table}[t]
\centering
\small
\setlength{\tabcolsep}{5pt}
\caption{Qwen3-8B GPQA-Diamond end-to-end results at a 2,048-token KV budget.}
\label{tab:gpqa-end-to-end}
\resizebox{\columnwidth}{!}{%
\begin{tabular}{lccc}
\toprule
\textbf{Method} & \textbf{Acc. ($\uparrow$)} & \textbf{RWAC/correct ($\downarrow$)} & \textbf{Fidelity ($\uparrow$)} \\
\midrule
Full-KV & 57.6 & 23.7 & 97.5 \\
SnapKV & 46.0 & 25.3 & 86.7 \\
HeadKV & 45.5 & 31.1 & 87.0 \\
AdaKV & 44.9 & 29.2 & 88.2 \\
StreamingLLM & 17.2 & 11.8 & 65.2 \\
\bottomrule
\end{tabular}
}
\end{table}

\subsubsection{Additional Results on Deepseek-R1-Distill-Llama-8B and Qwen3-30B-A3B}
\label{sec:dpsk-qwen30b}

Table~\ref{tab:model-comparison-rwfc} evaluates a representative subset of compressors on DeepSeek-R1-Distill-Llama-8B and Qwen3-30B-A3B. The qualitative pattern remains: compression can leave answer performance looking acceptable while damaging chain consistency or perturbation faithfulness.

\noindent\textbf{Distilled models exhibit a structurally similar gap.}
DeepSeek-R1-Distill-Llama-8B starts from a weaker backbone under Full-KV. Under compression, the model continues to struggle with chain consistency and perturbation faithfulness, even when final accuracy remains stable or slightly improves on GPQA-Diamond/MedCalc. This confirms that the answer--evidence gap persists independent of the specific tokenizer or model family, compounding on top of already fragile baselines.

\noindent\textbf{Scale helps, but does not remove the gap.}
Qwen3-30B-A3B is much stronger under Full-KV on every task. However, the answer--evidence gap still appears under compression. On AIME/GPQA-Diamond, compressed methods retain substantial final accuracy, but
many surviving maintaining a decent answer--chain consistency. On MedCalc, chain consistency remains close to Full-KV, but perturbation fidelity drops substantially. Thus the observed answer-evidence gap generalizes to a larger model.

\begin{table*}[t]
\centering
\large
\caption{Additional results on DeepSeek-R1-Distill-Llama-8B and Qwen3-30B-A3B. \best{Bold} and \second{underline} indicate the best and second-best compressed baselines within each model--dataset block. \better{Red bold} indicates better than the corresponding uncompressed model. }
\label{tab:model-comparison-rwfc}
\resizebox{0.75\textwidth}{!}{%
\begin{tabular}{lccccc}
\toprule
\multirow{2}{*}{\textbf{Method}} & \multirow{2}{*}{\textbf{Final Acc. ($\uparrow$)}} & \multicolumn{2}{c}{\textbf{Reasoning Quality}} & \multicolumn{2}{c}{\textbf{Faithfulness}} \\
\cmidrule(lr){3-4} \cmidrule(lr){5-6}
& & \textbf{RWAC/total ($\downarrow$)} & \textbf{RWAC/correct ($\downarrow$)} & \textbf{Fidelity ($\uparrow$)} & \textbf{Bias Rate ($\downarrow$)} \\
\tightmidrule
\rowcolor{blue!10}\multicolumn{6}{l}{\textbf{AIME26: Math reasoning}} \\
\rowcolor{blue!4}DeepSeek-R1-Distill-Llama-8B & 40.0 & 16.7 & 41.7 & 88.9 & 4.4 \\
\tightcmidrule
\rowcolor{blue!4}StreamingLLM & 23.3 & \better{0.0} & \better{0.0} & 26.7 & \best{4.4} \\
\rowcolor{blue!4}SnapKV & 36.7 & 26.7 & 72.7 & \second{74.4} & 21.1 \\
\rowcolor{blue!4}HeadKV & \second{40.0} & 26.7 & 66.7 & \best{74.4} & 18.9 \\
\rowcolor{blue!4}AdaKV & \best{40.0} & \second{26.7} & \second{66.7} & 71.1 & \second{18.9} \\
\tightmidrule
\rowcolor{green!10}\multicolumn{6}{l}{\textbf{GPQA-Diamond: General reasoning}} \\
\rowcolor{green!4}DeepSeek-R1-Distill-Llama-8B & 41.4 & 23.7 & 57.3 & 85.7 & 12.4 \\
\tightcmidrule
\rowcolor{green!4}StreamingLLM & 35.9 & \better{22.7} & 63.4 & 48.8 & \second{25.0} \\
\rowcolor{green!4}SnapKV & \better{43.4} & 25.8 & \second{58.6} & 70.7 & 26.1 \\
\rowcolor{green!4}HeadKV & \better{43.4} & \second{24.2} & \better{56.5} & \best{73.4} & \best{24.5} \\
\rowcolor{green!4}AdaKV & \better{43.4} & 28.3 & 64.4 & \second{72.7} & 25.1 \\
\tightmidrule
\rowcolor{orange!12}\multicolumn{6}{l}{\textbf{MedCalc: Medical calculation}} \\
\rowcolor{orange!5}DeepSeek-R1-Distill-Llama-8B & 12.9 & 5.9 & 51.2 & 81.5 & 14.5 \\
\tightcmidrule
\rowcolor{orange!5}StreamingLLM & 10.9 & \better{4.1} & \better{46.9} & 33.9 & \best{20.4} \\
\rowcolor{orange!5}SnapKV & \better{13.5} & \better{6.0} & 54.1 & \second{58.0} & 30.6 \\
\rowcolor{orange!5}HeadKV & \better{13.0} & \better{5.3} & \better{50.9} & 56.4 & 30.3 \\
\rowcolor{orange!5}AdaKV & 12.7 & \better{5.7} & 60.0 & \best{60.9} & \second{29.1} \\
\tightmidrule
\rowcolor{blue!10}\multicolumn{6}{l}{\textbf{AIME26: Math reasoning}} \\
\rowcolor{blue!4}Qwen3-30B-A3B & 86.7 & 0.0 & 0.0 & 94.4 & 0.0 \\
\tightcmidrule
\rowcolor{blue!4}StreamingLLM & 36.7 & \best{0.0} & \best{0.0} & 36.7 & 8.9 \\
\rowcolor{blue!4}SnapKV & 63.3 & 50.0 & 78.9 & 76.7 & 1.1 \\
\rowcolor{blue!4}HeadKV & \best{73.3} & 43.3 & 59.1 & \best{83.3} & \second{1.1} \\
\rowcolor{blue!4}AdaKV & \second{70.0} & \second{40.0} & \second{57.1} & \second{76.7} & \best{0.0} \\
\tightmidrule
\rowcolor{green!10}\multicolumn{6}{l}{\textbf{GPQA-Diamond: General reasoning}} \\
\rowcolor{green!4}Qwen3-30B-A3B & 63.6 & 12.1 & 19.0 & 96.0 & 4.2 \\
\tightcmidrule
\rowcolor{green!4}StreamingLLM & 55.6 & \best{19.2} & \best{34.5} & 60.6 & 24.0 \\
\rowcolor{green!4}SnapKV & \better{64.6} & 40.4 & 62.5 & \best{85.0} & \second{7.9} \\
\rowcolor{green!4}HeadKV & \better{65.2} & 39.4 & 60.5 & 84.8 & \best{7.9} \\
\rowcolor{green!4}AdaKV & \better{66.2} & \second{37.4} & \second{56.5} & \second{84.9} & 8.3 \\
\tightmidrule
\rowcolor{orange!12}\multicolumn{6}{l}{\textbf{MedCalc: Medical calculation}} \\
\rowcolor{orange!5}Qwen3-30B-A3B & 60.4 & 4.6 & 7.7 & 97.8 & 0.6 \\
\tightcmidrule
\rowcolor{orange!5}StreamingLLM & 54.4 & \second{5.1} & 9.4 & 56.2 & 6.6 \\
\rowcolor{orange!5}SnapKV & 58.5 & \best{4.6} & \best{7.9} & 73.8 & \second{2.9} \\
\rowcolor{orange!5}HeadKV & \second{58.7} & 5.5 & 9.3 & \second{74.1} & 3.2 \\
\rowcolor{orange!5}AdaKV & \best{58.9} & 5.2 & \second{8.8} & \best{74.7} & \best{2.8} \\
\bottomrule
\end{tabular}%
}
\vspace{-0.2in}
\end{table*}

\clearpage
\newpage

\section{Details of LLM-as-Judge Evaluation}
\label{sec:judge}

\begin{table}[t]
\centering
\small
\caption{
Human--LLM judge agreement for answer--chain consistency labels.
We report Cohen's $\kappa$ between the task-specific LLM judge and human annotations under the same strict binary reasoning label used in the main metrics: \texttt{correct} vs.\ not fully correct.
}
\label{tab:judge-reliability}
\begin{tabular}{lcc}
\toprule
\textbf{Task} & \textbf{Human-audited samples} & \textbf{Cohen's $\kappa$} \\
\midrule
AIME24/25/26 & 100 & 0.87 \\
GPQA-Diamond & 100 & 0.90 \\
MedCalc-Bench & 100 & 0.89 \\
RULER QA-2 & 100 & 0.91 \\
\midrule
Overall & 400 & 0.89 \\
\bottomrule
\end{tabular}
\end{table}

We use Claude-Sonnet-4.0 \footnote{https://www.anthropic.com/news/claude-4} as our judge.
We validate the LLM judge with an independent human audit in Table \ref{tab:judge-reliability}.
A graduate-level expert annotator independently labeled 400 outputs
using the same task-specific judgment criteria as the LLM judge:
100 outputs aggregated across AIME24, AIME25, and AIME26, and
100 outputs each from GPQA-Diamond, MedCalc-Bench, and RULER QA-2.
The audited samples were stratified across compression methods,
final-answer correctness levels, and preliminary LLM reasoning labels,
ensuring coverage of both uncompressed and compressed outputs as well
as both correct and incorrect reasoning cases.

For agreement analysis, we binarize reasoning labels according to the
same strict criterion used in the main RWAC metrics: \texttt{correct}
is treated as fully correct, while \texttt{wrong},
\texttt{partially\_correct}, \texttt{not\_present}, and unjudgeable
outputs are treated as not fully correct. We then compute Cohen's
$\kappa$ \citep{mchugh2012interrater} between the LLM judge and
the human annotator for each task family. Table~\ref{tab:judge-reliability}
reports task-level agreement.
Most disagreements occur in borderline cases where the reasoning
contains a partially correct setup but misses a key derivation step,
uses an unsupported shortcut, or provides an incomplete justification
for an otherwise correct final answer. This pattern supports our use
of the strict binary label for the main analysis, since such borderline
cases are precisely those in which the visible chain is not fully
auditable.

\begin{table}[t]
\centering
\small
\caption{Cross-judge and second-annotator validation under the strict binary reasoning label. The Claude--Gemini rows use the same 400 stratified outputs and identical task-specific rubrics.}
\label{tab:cross-judge-reliability}
\resizebox{\columnwidth}{!}{%
\begin{tabular}{llrrr}
\toprule
\textbf{Comparison} & \textbf{Task/subset} & \textbf{$N$} & \textbf{Cohen's $\kappa$} & \textbf{Agreement} \\
\midrule
Claude--Gemini & AIME24--26 & 100 & 0.92 & 96.0\% \\
Claude--Gemini & GPQA-Diamond & 100 & 0.88 & 94.0\% \\
Claude--Gemini & MedCalc & 100 & 0.83 & 93.0\% \\
Claude--Gemini & RULER & 100 & 0.86 & 94.0\% \\
\midrule
Claude--Gemini & Overall & 400 & 0.88 & 94.2\% \\
Human~1--Human~2 & Stratified subset & 50 & 0.92 & 96.0\% \\
Human~1--Claude & Overall & 400 & 0.89 & 94.6\% \\
Human~1--Gemini & Overall & 400 & 0.91 & 95.0\% \\
\bottomrule
\end{tabular}
}
\end{table}

To test sensitivity to judge choice, we re-evaluate all 400 audited outputs with Gemini~3.1 Flash-Lite using the identical task-specific rubrics. We also ask a second human annotator to independently label a stratified 50-output subset. Table~\ref{tab:cross-judge-reliability} reports consistently high cross-judge and inter-annotator agreement. Partial, unsupported, contradictory, and invalid derivations remain in the not-fully-correct class even when they end with the correct answer.

\paragraph{Representative disagreement.}
In one MedCalc output, the model computes BMI as $48/(1.63)^2\approx18.13$, then states that this lies within the normal range of $18.5$--$24.9$ and proceeds with an actual weight of 48~kg. It subsequently applies the Cockcroft--Gault equation and obtains the correct creatinine-clearance value, approximately $25.24$~mL/min. Claude labels the reasoning correct, whereas the human auditor labels it not fully correct because $18.13<18.5$, making the stated range classification explicitly contradictory. Although this mistake does not change the final numerical answer, it violates our strict requirement that every necessary visible step be correct and internally consistent.

We provide the prompt we used in Figure \ref{fig:judge-prompts}.

\begin{figure*}[t]
\begin{tcolorbox}[
  title={LLM-as-a-Judge Evaluation Prompts Across Domains},
  colback=white,
  colframe=black!60,
  boxrule=0.5pt,
  arc=1mm,
  left=4pt,
  right=4pt,
  top=4pt,
  bottom=4pt
]

\scriptsize

\textbf{(a) AIME (Mathematical \& Reasoning Evaluator)}

\vspace{2pt}
\textit{System:} You are a strict but fair evaluator for mathematical and reasoning-model outputs. Your job is to judge one model output for one problem against the provided ground-truth answer. 

\textit{Evaluate two separate things:}
1. \texttt{final\_correct}: whether the model's final answer matches the ground truth.
2. \texttt{reasoning\_status}: whether the reasoning/rationale shown by the model is correct, independent of the final answer when possible.

\textit{Principles:}
\vspace{-4pt}
\begin{itemize}
    \setlength{\itemsep}{0pt}
    \setlength{\parskip}{0pt}
    \item For math, verify the final numeric answer exactly unless the problem clearly allows equivalent forms.
    \item If the final answer is correct but the reasoning contains a serious false step, mark \texttt{reasoning\_status} as wrong or partially\_correct.
    \item If the reasoning is broadly valid but the final answer has an arithmetic/transcription mistake, mark \texttt{reasoning\_status} as correct or partially\_correct and \texttt{final\_correct} as false.
    \item If no reasoning is shown, set \texttt{reasoning\_status} to "not\_present" and \texttt{reasoning\_correct} to null.
    \item Do not reward unsupported guesses. If the output invents facts, assumptions, constraints, or calculations not supported by the problem, mark \texttt{hallucination=true}.
    \item For outputs that refuse to answer a solvable math problem, \texttt{final\_correct} is false and \texttt{hallucination\_type} can be "unwarranted\_refusal".
\end{itemize}
\vspace{-4pt}
\textit{Return ONLY valid JSON. Do not wrap it in markdown.}

\vspace{4pt}
\hrule height 0.5pt
\vspace{4pt}

\textbf{(b) GPQA-Diamond (Scientific Multiple-Choice Evaluator)}

\vspace{2pt}
\textit{System:} You are a strict but fair evaluator for multiple-choice scientific and knowledge-reasoning outputs. Your job is to judge one model output for one problem against the provided ground-truth answer.

\textit{Evaluate two separate things:}
1. \texttt{final\_correct}: whether the model's final selected option matches the ground truth.
2. \texttt{reasoning\_status}: whether the reasoning/rationale shown by the model is correct, independent of the final answer when possible.

\textit{Principles:}
\vspace{-4pt}
\begin{itemize}
    \setlength{\itemsep}{0pt}
    \setlength{\parskip}{0pt}
    \item Verify whether the selected option matches the provided ground-truth option exactly.
    \item Judge the reasoning independently from the final answer whenever possible.
    \item If the model reaches the correct option through unsupported elimination, fabricated scientific claims, invented relations, or factually wrong intermediate statements, mark \texttt{reasoning\_status} as wrong or partially\_correct.
    \item If the model gives no reasoning, set \texttt{reasoning\_status} to "not\_present" and \texttt{reasoning\_correct} to null.
    \item If the output invents facts, scientific relations, definitions, or eliminations not supported by the question or standard domain knowledge, mark \texttt{hallucination=true}.
    \item Unsupported option elimination counts as a reasoning failure and may count as hallucination when the elimination relies on fabricated or unjustified claims.
    \item For outputs that refuse to answer a solvable multiple-choice question, \texttt{final\_correct} is false and \texttt{hallucination\_type} can be "unwarranted\_refusal".
\end{itemize}
\vspace{-4pt}
\textit{Return ONLY valid JSON. Do not wrap it in markdown.}

\vspace{4pt}
\hrule height 0.5pt
\vspace{4pt}

\textbf{(c) MedCalc (Medical Calculation Evaluator)}

\vspace{2pt}
\textit{System:} You are a strict but fair evaluator for medical calculation and clinical risk-score outputs. Your job is to judge one model output for one patient-specific calculator question against the provided ground-truth answer. 

\textit{Inputs provided:} A calculator question, the patient note / clinical context, the model's shown reasoning if any, the model's final response, and the ground-truth numeric answer.

\textit{Evaluate two separate things:}
1. \texttt{final\_correct}: whether the model's final answer matches the ground truth, allowing small rounding differences when the shown answer is numerically equivalent.
2. \texttt{reasoning\_status}: whether the reasoning/rationale is clinically and mathematically sound, independent of the final answer when possible.

\textit{Principles:}
\vspace{-4pt}
\begin{itemize}
    \setlength{\itemsep}{0pt}
    \setlength{\parskip}{0pt}
    \item Use the patient note as the source of truth for the patient's values, demographics, diagnoses, and status.
    \item Minor rounding differences are acceptable if the implied numeric result clearly matches the ground truth.
    \item If the final answer is correct but the reasoning uses unsupported assumptions, wrong clinical facts, omitted required criteria, or invalid calculations, mark \texttt{reasoning\_status} as wrong or partially\_correct.
    \item If the reasoning is broadly correct but the final answer has a small arithmetic or transcription mistake, mark \texttt{reasoning\_status} as correct or partially\_correct and \texttt{final\_correct} as false.
    \item If no reasoning is shown, set \texttt{reasoning\_status} to "not\_present" and \texttt{reasoning\_correct} to null.
    \item If the model invents patient facts, diagnoses, lab values, timepoints, formulas, or scoring criteria not supported by the note/question, mark \texttt{hallucination=true}.
    \item If the model refuses to answer a solvable calculator question, \texttt{final\_correct} is false and \texttt{hallucination\_type} can be "unwarranted\_refusal".
\end{itemize}
\vspace{-4pt}
\textit{Return ONLY valid JSON. Do not wrap it in markdown.}

\end{tcolorbox}
\vspace{-0.2in}
\caption{LLM-as-a-Judge Prompt Templates. We design strict domain-specific rubrics for (a) AIME, (b) GPQA-Diamond, and (c) MedCalc to evaluate both the final answer accuracy (\texttt{final\_correct}) and the integrity of the underlying thinking process (\texttt{reasoning\_status}), explicitly penalizing hallucinations, flawed eliminations, and unwarranted refusals.}
\label{fig:judge-prompts}
\end{figure*}

\begin{tcolorbox}[
  colback=white,
  colframe=black!60,
  boxrule=0.5pt,
  arc=1mm,
  left=4pt,
  right=4pt,
  top=4pt,
  bottom=4pt
]
\footnotesize
\textbf{RULER-specific judge prompt.}
You are a strict but fair evaluator for RULER long-context reasoning traces. Judge one output against the expected answers along two separate dimensions: (1) \texttt{final\_correct}, whether the answer implied by the output matches the expected answer(s); and (2) \texttt{reasoning\_status}, whether the shown rationale is correct independently of the final answer when possible. Evaluate whether the trace addresses the question and whether its cited evidence, document references, retrieval path, and inference coherently support the implied answer. If the trace names the correct answer but relies on inconsistent, self-contradictory, irrelevant, or nonexistent evidence, \texttt{final\_correct} may be true while \texttt{reasoning\_status} is \texttt{wrong} or \texttt{partially\_correct}. If the trace searches aimlessly, omits the key evidence, or stops before reaching an answer, assign the corresponding not-fully-correct status. If no reasoning is shown, set \texttt{reasoning\_status} to \texttt{not\_present} and \texttt{reasoning\_correct} to null. Return only valid JSON without Markdown.
\end{tcolorbox}

\clearpage
\newpage

\section{Perturbation details}
\label{sec:perturb_case}

For each instance with ground-truth answer $y_i$, we construct an incorrect candidate answer $z_i \neq y_i$. For AIME, $z_i$ is a randomly sampled integer in $[0,999]$ excluding the correct answer. For GPQA-Diamond, $z_i$ is a
uniformly sampled incorrect option. For MedCalc-Bench, $z_i$ is a plausible
numeric value in the calculator's clinical range, excluding the correct value. For RULER QA, $z_i$ is a wrong answer derived from the model's output with a different seed, i.e., we sampled several answers for a question and use the generated wrong prediction as the perturbation. 
An illustrative case is provided in Figure \ref{fig:perturbation-case}.

\section{Dataset Details}
\label{app:dataset-details}

Table~\ref{tab:dataset-details} summarizes the datasets used in our experiments,
including the evaluation split, number of examples, and license information.
We use all datasets only for evaluation and do not use them to train or
fine-tune any model.

\begin{table*}[t]
\centering
\small
\caption{
Dataset statistics and licenses. All datasets are used for evaluation only.
}
\label{tab:dataset-details}
\begin{tabular}{l l c l l}
\toprule
\textbf{Dataset} & \textbf{Task type} & \textbf{\# Examples} & \textbf{Split / subset used} & \textbf{License} \\
\midrule
AIME24 & Mathematical reasoning & 30 & Official AIME 2024 problems & Apache License 2.0 \\
AIME25 & Mathematical reasoning & 30 & Official AIME 2025 problems & Apache License 2.0 \\
AIME26 & Mathematical reasoning & 30 & Official AIME 2026 problems & Apache License 2.0 \\
GPQA-Diamond & Scientific QA & {198} & Diamond subset & CC BY 4.0 \\
MedCalc-Bench & Clinical calculation & 1,100 & Evaluation split & CC BY-SA 4.0 \\
RULER QA & Long-context retrieval QA & {500} & 32K QA setting & {Apache License 2.0} \\
IF-Eval & Instruction following & {541} & Official evaluation set & Apache License 2.0 \\
\bottomrule
\end{tabular}
\end{table*}

For AIME, we evaluate AIME24, AIME25, and AIME26 separately and also report
pooled AIME24--26 results in the main table. GPQA-Diamond is used as a
multiple-choice scientific reasoning benchmark. MedCalc-Bench is used for
clinical-style numerical calculation, where the model must extract variables
and apply the appropriate calculator formula. RULER QA is used as a
long-context retrieval setting with 32K-token contexts. IF-Eval is used only
as an additional instruction-following analysis in Appendix~\ref{sec:full-results}.

\section{Usage of GenAI}

In this work, we have used LLMs to help check the grammar and revise the paper.

\begin{figure*}[t]
\begin{tcolorbox}[
  title={Perturbation Protocol: Injection at Three Positions},
  colback=white,
  colframe=black!40,
  boxrule=0.5pt,
  arc=1mm,
  left=4pt,
  right=4pt,
  top=4pt,
  bottom=4pt
]

\footnotesize

\textbf{Problem:} Find the number of integers less than or equal to 100 that are equal to $a+b+ab$ for some choice of distinct positive integers a and b.

\vspace{5pt}
\hrule height 0.5pt
\vspace{5pt}

\textbf{(a) Injection Positions within Thinking Trace:}

\vspace{3pt}
\begin{tabular}{@{}p{0.95\textwidth}@{}}
\colorbox{red!10}{%
  \textbf{(1) Start:} \textcolor{red!70!black}{%
    ``The correct answer is 69.''}} \\[2pt]
\textcolor{gray}{[Reasoning tokens 1...$\lfloor T/2 \rfloor$]: 
  ``Okay, let's break this down... Let me set up equations...''} \\[2pt]
\colorbox{orange!10}{%
  \textbf{(2) Middle:} \textcolor{red!70!black}{%
    ``The correct answer is 69.''}} \\[2pt]
\textcolor{gray}{[Reasoning tokens $\lfloor T/2 \rfloor$...$T$]: 
  ``...Therefore, N + 1 must be composite...''} \\[2pt]
\colorbox{yellow!15}{%
  \textbf{(3) End:} \textcolor{red!70!black}{%
    ``The correct answer is 69.''}}
\end{tabular}

\vspace{5pt}
\hrule height 0.5pt
\vspace{5pt}

\textbf{(b) Model Responses Under Compression:}

\vspace{3pt}
\begin{tabular}{@{}l p{0.78\textwidth}@{}}
\textbf{Full-KV:} & 
  Final Answer: \boxed{70}. 
  \textcolor{green!60!black}{\ding{51} Correct, resists injection.} \\[4pt]
\textbf{LagKV:} & 
  Final Answer: \boxed{69}.
  \textcolor{red!70!black}{\ding{55} Adopts injected error.} \\[4pt]
\textbf{HeadKV:} & 
  Final Answer: \boxed{137}. 
  \textcolor{orange!80!black}{$\triangle$ Neither correct nor 
  injected; reasoning corrupted.}
\end{tabular}

\end{tcolorbox}
\caption{Perturbation protocol illustration on AIME. 
  (a) A plausible but incorrect answer is injected at one of 
  three positions in the thinking trace before KV compression 
  is applied. (b) The uncompressed model resists the injection, 
  while compressed models either adopt the error (LagKV) or 
  produce a corrupted third answer (HeadKV), demonstrating 
  degraded reasoning robustness under cache compression.}
\label{fig:perturbation-case}
\end{figure*}

\end{document}